\documentclass{article}

    \PassOptionsToPackage{numbers, compress}{natbib}
    \usepackage[preprint]{neurips_2023}

\usepackage[utf8]{inputenc} % allow utf-8 input
\usepackage[T1]{fontenc}    % use 8-bit T1 fonts
\usepackage{hyperref}       % hyperlinks
\usepackage{url}            % simple URL typesetting
\usepackage{booktabs}       % professional-quality tables
\usepackage{amsfonts}       % blackboard math symbols
\usepackage{nicefrac}       % compact symbols for 1/2, etc.
\usepackage{microtype}      % microtypography
\usepackage{xcolor}         % colors
\usepackage{graphicx}
\usepackage{amsmath}
\usepackage{amssymb}
\usepackage{mathtools}
\usepackage{amsthm}
\usepackage{wrapfig}
\usepackage{float}

\usepackage{dsfont}
\usepackage{multirow}
\usepackage{float,paralist}
\usepackage{transparent}
\usepackage{stfloats}

\usepackage{algorithm}
\usepackage{algorithmic}

\DeclarePairedDelimiterX{\inner}[1]{\langle}{\rangle}{#1}

\newcommand{\T}{\top}

\newcommand{\vv}{\mathbf{v}}

\newcommand{\mA}{\mathbf{A}}
\newcommand{\mB}{\mathbf{B}}
\newcommand{\mC}{\mathbf{C}}

\newcommand{\mI}{\mathbf{I}}

\newcommand{\mK}{\mathbf{K}}

\newcommand{\mM}{\mathbf{M}}

\newcommand{\mO}{\mathbf{O}}
\newcommand{\mP}{\mathbf{P}}
\newcommand{\mQ}{\mathbf{Q}}

\newcommand{\mS}{\mathbf{S}}

\newcommand{\mV}{\mathbf{V}}
\newcommand{\mW}{\mathbf{W}}
\newcommand{\mX}{\mathbf{X}}
\newcommand{\mY}{\mathbf{Y}}

\newcommand{\mPi}{\mathbf{\Pi}}

\title{IM-Unpack: Training and Inference with Arbitrarily Low Precision Integers}

\theoremstyle{plain}
\newtheorem{theorem}{Theorem}[section]

\theoremstyle{definition}

\theoremstyle{remark}
\newtheorem{remark}[theorem]{Remark}

\author{%
  Zhanpeng Zeng $\quad$
  Karthikeyan Sankaralingam $\quad$
  Vikas Singh\\ 
University of Wisconsin--Madison \\
\texttt{zzeng38@wisc.edu} $\quad$
\texttt{karu@cs.wisc.edu} $\quad$
\texttt{vsingh@biostat.wisc.edu} \\
}

\begin{document}

\maketitle

\begin{abstract}

% {\color{red}
% Transformers have dominated the landscape of language and vision. 
% The fast growth of model sizes for Transformer based large language models and vision models put much heavier demands on hardware and software to improve the speed for training and serving these large models. 

% Many works seek to use quantization to use lower precision computation to reduce the computation and memory demands of these models. 
GEneral Matrix Multiply (GEMM) is a central operation in deep learning and corresponds to the largest chunk of the compute footprint. Therefore, improving its efficiency is an active topic of ongoing research. A popular strategy is the use of low bit-width integers to approximate the original entries in a matrix. This allows efficiency gains, but often requires sophisticated techniques to control the rounding error incurred. In this work, we first verify/check that when the low bit-width restriction is removed, for a variety of Transformer-based models, whether integers are sufficient for all GEMMs need -- for {\em both} training and inference stages, and can achieve parity with floating point counterparts. No sophisticated techniques are needed. We find that while a large majority of entries in matrices (encountered in such models) can be easily represented by {\em low} bit-width integers, the existence of a few heavy hitter entries make it difficult to achieve efficiency gains via the exclusive use of low bit-width GEMMs alone. To address this issue, we develop a simple algorithm, Integer Matrix Unpacking (IM-Unpack), to {\em unpack} a matrix with large integer entries into a larger matrix whose entries all lie within the representable range of arbitrarily low bit-width integers. This allows {\em equivalence} with the original GEMM, i.e., the exact result can be obtained using purely low bit-width integer GEMMs. This comes at the cost of additional operations -- we show that for many popular models, this overhead is quite small. 
% }

% Transformer based models and represents the heaviest computation in both forward and backward pass of a Transformer compared to other operations such as non-linear activation and normalization. 
% In this work, we first verified that simply using  to uniformly quantize matrices involved in each GEMM in forward and backward pass of a model surprisingly works well. For the cases of Transformer based language models and Vision Transformers, we observed no or little performance degradation for both training and inference without sophisticated training or finetuning techniques. 
% However, the outliers presented during quantization makes low precision integer computation insufficient and requires higher precision to ensure the outliers fall into the representable range, resulting in waste of compute resource. 

\end{abstract}

\section{Introduction}

% The Transformer \citep{NIPS2017_3f5ee243} has become the fundamental architecture for language modeling and vision applications. Transformer based large language models (LLMs) have dominated the landscape of natural language processing (NLP) tasks \citep{touvron2023llama, openai2023gpt4, workshop2023bloom}, and Vision Transformers (ViT) have became popular choice for classification \citep{vit}, object detection \citep{detr}, generation \citep{esser2021taming, Chang_2022_CVPR}, and more \citep{cheng2022masked, ranftl2021vision}. 
% Recent results \citep{vit-22b, zhai2022scaling, touvron2023llama, openai2023gpt4} have demonstrated favorable results on larger model size: we could expect better performance with larger model size and larger dataset. 
% While these large models provide remarkable performance, the fast growing model size of these large Transformers puts heavy demands on compute and memory demands. Serving these model for inference requires expensive hardware and energy. Using these models on consumer-grade machines can be slow to be practically used or even infeasible. 
% Many proposed strategies specifically targeted to improve the general efficiency of these Transformers \citep{dao2022flashattention, beltagy2020longformer, zeng2023vcc, Lan2020ALBERT}. There are also techniques for efficient inference including distillation \citep{hinton2015distilling}, pruning \citep{lee2018snip, prune1} and quantization \citep{NEURIPS2019_c0a62e13, nagel2019data, I-BERT, dettmers2022gptint, ivit}. 

Calculating the 
product of two matrices using GEneral Matrix Multiply (GEMM) is one of the 
most widely used operations in modern machine learning. 
Given matrices $\mA$ and $\mB$ of size $n \times d$ and $h \times d$ respectively, the output of a GEMM is calculated as
\begin{equation}
\mC = \mA \mB^\T
\label{eq:gemm}
\end{equation}
Choosing the appropriate numerical precision or data type (FP32, FP16, or BF16) for GEMM is often important, and hinges on several factors including the specific application,  characteristics of the data, model architecture, as well as numerical behavior such as convergence. 
This choice affects compute and memory efficiency most directly, since 
a disproportionately large chunk of the compute footprint of a model 
involves the GEMM operator. 
%GEMM corresponds to the largest chunk of the compute footprint of the model. 
% Indeed, enormous gains in compute bandwidth 
% of GPU chips (from Pascal to Volta to Ampere) are attributed to specialized accelerators for GEMM. But even aside from  
% hardware improvements, efficiency gains for GEMM can 
% also come from software optimization. 
A good example is 
the large improvement in latency and memory achieved via low bit-width GEMM, 
% (only involves some hardware support for the operation), 
and made possible due to 
extensive ongoing work on quantization (to low bit-width data types) and low-precision training 
\citep{NEURIPS2019_c0a62e13, nagel2019data, I-BERT, dettmers2022gptint, ivit, xiao2023smoothquant, dettmers2022gptint, liu2023llmqat, llmfp4, lin2022fqvit, ivit, PTQ4ViT_arixv2022, apq-vit, li2023repq, wang2018training, wu2018training, zhu2020towards, wortsman2023stable}. 
Integer quantization is being actively pursued for inference efficiency, and the use of {\em low bit-width} integers is universal to deliver the efficiency gains. 
However, this strategy often incurs large rounding errors when representing all matrix entries as low bit-width integers, and explains the drop in performance and thereby, a need for error correction techniques \citep{frantar2023optq, xiao2023smoothquant, chee2023quip}. 
So how much of the performance degradation is due 
to (a) rounding to integers versus (b) restricting to low bit-width integers?
To answer this question, it appears worthwhile to check 
whether integer GEMMs will achieve parity without sophisticated techniques
(for the inference stage, and more aspirationally, for training) for popular models if we do {\em not} restrict to low bit-width integers. 

\begin{figure*}[!tb]
\centering
% \vspace{-0.05in}
\includegraphics[width=0.99\textwidth]{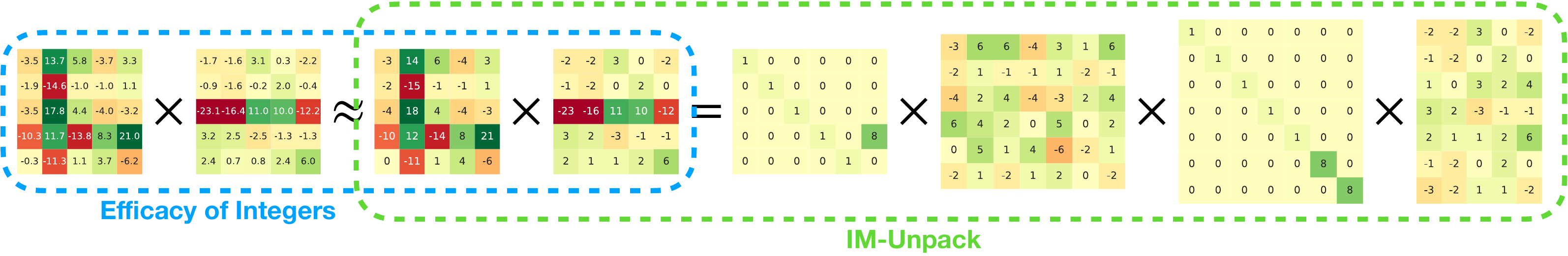}
% \vspace{-0.1in}
\caption{Overall Illustration. We verify the \textbf{Efficacy of Integers (Contribution 1)} in \S\ref{sec:integer}, but note that the integer matrices contain heavy hitters (\S\ref{sec:issue_low_precision_int}). Then, we describe our proposed algorithm, \textbf{IM-Unpack (Contribution 2)}, to resolve these heavy hitters in \S\ref{sec:imunpack}. }
\label{fig:main_figure}
\vspace{-0.15in}
\end{figure*}

%This suggests  for efficiency, then, . 
{\bf Overview.} The starting point of our work is to first 
experimentally verify that the 
aforementioned hypothesis -- that 
% ({\em not} low bit-width)
integer GEMM may work -- is true (see \S\ref{sec:integer}). 
But by itself, this finding offers no value proposition 
for efficiency. 
Still, this experiment is useful for the following reason. For a particular 
class of models (e.g., Transformers), we can easily contrast the corresponding input matrices $\mA$ and $\mB$ between (a) integer GEMM and (b) low bit-width integer GEMM and probe if any meaningful structure can be exploited.  
While there is a clear difference in the {\em outputs} 
of (a) integer GEMM versus (b) low bit-width integer GEMM, 
we find that 
%allows probing for the structures of $\mA$ and $\mB$ to efficiency opportunities. 
{\em a large majority} of entries of $\mA$ and $\mB$ can be well-represented using low bit-width integers -- and the difference in the outputs 
can be entirely attributed to a few {\em heavy hitter} entries in $\mA$ and $\mB$, that cannot be represented using low bit-width integers. 
%usually contains  that cannot be represented using low bit-width integers, which requires special treatments. 
Other works have also run into this issue of ``outliers'' 
and suggest using high precision \citep{dettmers2022gptint} or a separate quantization for these entries \citep{PTQ4ViT_arixv2022, xiao2023smoothquant}.
%Fortunately, most of these heavy hitters usually concentrate on some rows or columns enabling different strategies for handling these entries including use high precision for these entries \citep{dettmers2022gptint} or use separate quantization for these entries \citep{PTQ4ViT_arixv2022, xiao2023smoothquant}. However, we note that these might requires GEMM hardware support for different precision or might lower the performance. 
%Alternatively, we demonstrate that these heavy hitters might not be that different from the remaining entries. 
Driven by the simple observation that we can represent a large integer by a series of 
% (or in a basis of)
smaller integers, our algorithm, 
Integer Matrix Unpack (IM-Unpack), enables unpacking 
% heavy hitters
any integer
% {\em recursively}
into a series of 
low-bit integers. The key outcome is that 
the calculation can be carried 
out entirely using low bit-width integer arithmetic and thus unifies the computation needed for heavy hitters and the remaining entries (which were already amenable to low-bit integer arithmetic).
Specifically, 
IM-Unpack 
% recursively
unpacks an integer matrix 
such that all values of the unpacked matrices always stay within the representable range of low bit-width integers (bit-width can be chosen arbitrarily). 
We obtain the exact result of the original integer GEMM using purely low bit-width integer GEMMs. 
% {\color{red} 
Since the bit-width of integer arithmetic is independent of the actual range of the original matrices, the construction will greatly simplify the hardware/compiler support by only needing support for {\em one} specific bit-width. The overall structure/contributions of this paper is shown in Fig. \ref{fig:main_figure}. 

{\bf Notations.}
To simplify the presentation, we will narrow the scope of our discussion exclusively to Transformer-based models. We first define notations for all relevant GEMMs. 
For the linear layer, let the 
input activation and parameter matrix be $\mX$ and $\mW$. Let the query, key, value matrices involved in self-attention computation be $\mQ, \mK, \mV$. Below, we itemize 
all GEMMs used in a Transformer model:
\begin{equation}
\begin{split}
\mY = \mX \mW^\T \quad \mP = \mQ \mK^\T \quad \mO = \mM \mV
\end{split}
\end{equation}
where $\mM$ is the attention score between $\mQ$ and $\mK$ defined as $\mM = \text{softmax}(\mP)$ (omitting scaling factors). 
Now, given the gradient for $\mY, \mP, \mO$ denoted as $\nabla_{\mY}, \nabla_{\mP}, \nabla_{\mO}$, the other gradients are calculated via GEMMs as well: 
\begin{equation}
\begin{split}
\nabla_{\mX} = \nabla_{\mY} \mW \quad \nabla_{\mQ} &= \nabla_{\mP} \mK \quad \nabla_{\mM} = \nabla_{\mO} \mV^\T \\
\nabla_{\mW} = \nabla_{\mY}^\T \mX \ \quad \nabla_{\mK} &= \nabla_{\mP}^\T \mQ \quad \ \nabla_{\mV} = \mM^\T \nabla_{\mO} \\
\end{split}
\end{equation}
% Numerous efforts have been devoted to improve the efficiency of the above calculation. 
These notations will help refer to each type of GEMM later.

\section{Round to Nearest: What do we lose? }

\begin{table}[!tb]
% \vspace{-0.09in}
\caption{Inference: Comparison on LLaMA-7B zero-shot performance and ViT ImageNet classification when using quantized computations in all linear layers. HS: HellaSwag, WG: WinoGrande. The super-script $^\ddag$ indicates that LLM.int8() uses mixed-precision (INT8+FP16) to process outliers using FP16. }
\label{tab:quantize_linear}
\vspace{-0.1in}
\begin{center}
\begin{small}
\setlength{\tabcolsep}{2.6pt}
\begin{tabular}{clllcccccc}
\toprule
\parbox[t]{3mm}{\multirow{12}{*}{\rotatebox[origin=c]{90}{LLaMA-7B}}} & Method & $\beta$ & Type & ARC-c & ARC-e & BoolQ & HS & PIQA & WG \\
\cmidrule{2-10}
& Full-Precision & - & BF16 & 43.1 & 76.3 & 77.8 & 57.2 & 78.0 & 68.8 \\
\cmidrule{2-10}
& LLM.int8() & - & INT8$^\ddag$ & 43.8 & 75.5 & 77.8 & 57.4 & 77.6 & 68.7 \\
& SmoothQuant & - & INT8 & 37.4 & 74.4 & 74.0 & 55.0 & 77.5 & 69.6 \\
& LLM-QAT & - & INT4 & 30.2 & 50.3 & 63.5 & 55.6 & 64.3 & 52.9 \\
& LLM-FP4 & - & FP4 & 33.6 & 65.9 & 64.2 & 47.8 & 73.5 & 63.7 \\
\cmidrule{2-10}
& RTN & 5 & INT & 39.3 & 72.8 & 69.9 & 53.4 & 74.9 & 66.4 \\
&  & 7 & INT & 42.6 & 73.9 & 72.3 & 55.9 & 77.0 & 67.4 \\
 % & 9 & INT & 43.0 & 74.5 & 75.7 & 56.4 & 77.4 & 69.4 \\
&  & 11 & INT & 43.9 & 76.1 & 77.3 & 56.3 & 77.3 & 69.3 \\
 % & 13 & INT & 44.2 & 75.8 & 77.4 & 56.3 & 77.8 & 69.0 \\
&  & 15 & INT & 43.0 & 75.7 & 77.5 & 57.0 & 78.0 & 69.2 \\
&  & 31 & INT & 42.7 & 76.1 & 76.1 & 57.3 & 77.3 & 69.3 \\
\cmidrule[0.75pt]{1-10}
\end{tabular}
\setlength{\tabcolsep}{5pt}
\begin{tabular}{clllccccc}
% \toprule
\parbox[t]{3mm}{\multirow{6}{*}{\rotatebox[origin=c]{90}{ViT}}} & Method & $\beta$ & Type & Tiny & Small & Base & Large & Huge \\
\cmidrule{2-9}
& Full-Precision & - & FP32 & 75.5 & 81.4 & 85.1 & 85.8 & 87.6 \\
\cmidrule{2-9}
\cmidrule{2-9}
& RTN & 5 & INT & 3.9 & 36.9 & 78.7 & 83.6 & 85.3 \\
&  & 7 & INT & 41.0 & 70.9 & 82.8 & 84.9 & 86.7 \\
&  & 15 & INT & 71.4 & 79.8 & 84.6 & 85.6 & 87.5 \\
\bottomrule
\end{tabular}
\end{small}
\end{center}
\vspace{-0.1in}
\end{table}

\begin{table}[!tb]
% \vspace{-0.09in}
\caption{Inference: Comparison on LLaMA-7B and ViT when quantize computation in all GEMMs. $^*$: PTQ4ViT uses a twin uniform quantization so GEMMs cannot be performed on INT6 directly and requires some modifications. }
\label{tab:quantized_matmul}
\vspace{-0.1in}
\begin{center}
\begin{small}
\setlength{\tabcolsep}{3.07pt}
\begin{tabular}{clllcccccc}
\toprule
\parbox[t]{3mm}{\multirow{8}{*}{\rotatebox[origin=c]{90}{LLaMA-7B}}} & Method & $\beta$ & Type & ARC-c & ARC-e & BoolQ & HS & PIQA & WG \\
\cmidrule{2-10}
& Full-Precision & - & BF16 & 43.1 & 76.3 & 77.8 & 57.2 & 78.0 & 68.8 \\
\cmidrule{2-10}
& RTN & 5 & INT & 23.5 & 34.3 & 54.8 & 32.5 & 57.6 & 49.7 \\
& & 7 & INT & 34.2 & 64.0 & 64.6 & 50.1 & 70.3 & 61.2 \\
 % & 9 & INT & 37.4 & 68.2 & 69.3 & 53.4 & 72.2 & 62.7 \\
& & 11 & INT & 41.6 & 72.4 & 68.7 & 55.1 & 75.4 & 65.1 \\
 % & 13 & INT & 40.4 & 73.1 & 74.1 & 55.6 & 76.1 & 65.7 \\
& & 15 & INT & 44.0 & 75.0 & 74.6 & 56.4 & 77.0 & 66.3 \\
& & 31 & INT & 43.4 & 75.8 & 76.8 & 57.5 & 77.4 & 68.4 \\
% \bottomrule
\cmidrule[0.75pt]{1-10}
\end{tabular}
\setlength{\tabcolsep}{5pt}
\begin{tabular}{clllccccc}
% \toprule
\parbox[t]{3mm}{\multirow{11}{*}{\rotatebox[origin=c]{90}{ViT}}} & Method & $\beta$ & Type & Tiny & Small & Base & Large & Huge \\
\cmidrule{2-9}
& Full-Precision & - & FP32 & 75.5 & 81.4 & 85.1 & 85.8 & 87.6 \\
\cmidrule{2-9}
& FQ-ViT & - & INT8 & - & - & 83.3 & 85.0 & - \\
& I-ViT & - & INT8 & - & 81.3 & 84.8 & - & - \\
& PTQ4ViT & - & INT6$^*$ & 66.7 & 78.3 & 82.9 & 84.9 & 86.6 \\
& APQ-ViT & - & INT4 & 17.6 & 48.0 & 41.4 & - & - \\
& RepQ-ViT & - & INT4 & - & 65.1 & 68.5 & - & - \\
\cmidrule{2-9}
& RTN & 5 & INT & 3.5 & 28.5 & 76.9 & 83.2 & 84.9 \\
& & 7 & INT & 39.0 & 69.9 & 82.1 & 84.7 & 86.5 \\
& & 15 & INT & 71.1 & 79.8 & 84.5 & 85.6 & 87.5 \\
\bottomrule
\end{tabular}
\end{small}
\end{center}
\vspace{-0.1in}
\end{table}

% \textcolor{red}{more}

% \subsection{Simply Rounding to Nearest Numbers}

\label{sec:integer}

%In this section, 
%we verify the hypothesis that integer GEMMs should be sufficient for training and inference if we do not restrict to low bit-width integers. 
Let us start by using the simplest Rounding To Nearest (RTN) to map FP to integers, and check 
the extent to which integer GEMMs work satisfactorily for 
both training and inference, if we do not restrict to low bit-width integers. 
Specifically, for matrix $\mA$, 
\textbf{all} entries of $\mA$ are quantized via
% the quantized $\mA_q$ is %obtained via
\begin{equation}
\mA_q = \text{round}(0.5 \beta / \alpha_{p}(\mA) \mA)
\label{eq:quantized_A}
\end{equation}
where $\alpha_{p}(\mA)$ gives the $p$-th percentile (see \S\ref{ap:percentiles}) based on the magnitude of entries in $\mA$, i.e., $p\%$ of entries in $\mA$ fall in the interval $[- \alpha_{p}(\mA), \alpha_{p}(\mA)]$. We only need $\alpha_{p}(\mA)$ as a meaningful estimate of the approximate range of values, and so we set $p = 95\%$ for all experiments except 
a few cases discussed explicitly. 
The hyperparameter $\beta$ is the number of distinct integers that we want to use to encode values that are within $[- \alpha_{p}(\mA), \alpha_{p}(\mA)]$. 
Then, after quantization, the GEMM for the original matrices can be approximated (because we incur a rounding error) in the quantized domain using integer GEMMs. The approximated GEMM is computed using the quantized $\mA$ and $\mB$: 
\begin{equation}
\mC \approx \frac{ \alpha_{p}(\mA) \alpha_{p}(\mB) }{(0.5 \beta)^2} \mA_q \mB_q^\T
\label{eq:quantized_gemm}
\end{equation}
The scaling factor in \eqref{eq:quantized_gemm} is used to undo the scaling in  \eqref{eq:quantized_A}. Here,  $\mA_q \mB_q^\T$ is an integer GEMM, as desired. 
For notational simplicity, if clear from context, we will drop the $q$ subscript from $\mA$ and $\mB$. 
%In the \S\ref{sec:inference} and \S\ref{sec:training}, we evaluate the efficacy of integer GEMMs as a replacement to floating point GEMMs by evaluating how well simple RTNI works on inference and training. 

% \textbf{Step-1: Floating Point To Integer.} 
% We consider the quantization as a two step process: converting a floating point matrix to an integer matrix and representing the integer values using certain bit widths. 

\subsection{Efficacy of Integers: Inference}
\label{sec:inference}

A majority of the literature on quantized low precision calculations focuses on inference efficiency  \citep{frantar2023optq, chee2023quip, llmfp4, PTQ4ViT_arixv2022, frantar2023optq, chee2023quip, llmfp4, PTQ4ViT_arixv2022, lin2022fqvit, ivit, PTQ4ViT_arixv2022, apq-vit, li2023repq}. Here, given a trained model, quantization seeks to reduce the precision of parameters and input activations to low precision. This allows faster low precision arithmetic for compute efficiency while maintaining model performance. So, we first evaluate how well RTN preserves model performance compared to baselines in this inference regime.
% {\color{red}(we quantize all values) when quantizing the majority of values in matrices to integers}. 
See \S\ref{ap:quantize_parameters} for results for quantizing parameters for memory saving and see \S\ref{ap:more_inference_experiments} for other experiments. 
Most quantization schemes for LLMs focus on quantizing GEMMs in Linear layers, while quantization methods for Vision Transformers are more ambitious and quantize {\em all} GEMMs in a Transformer. We follow this convention for baselines, but present all variants for RTN. 

\begin{wrapfigure}{R}{0.5\textwidth}
\centering
\vspace{-0.1in}
\includegraphics[width=0.5\textwidth]{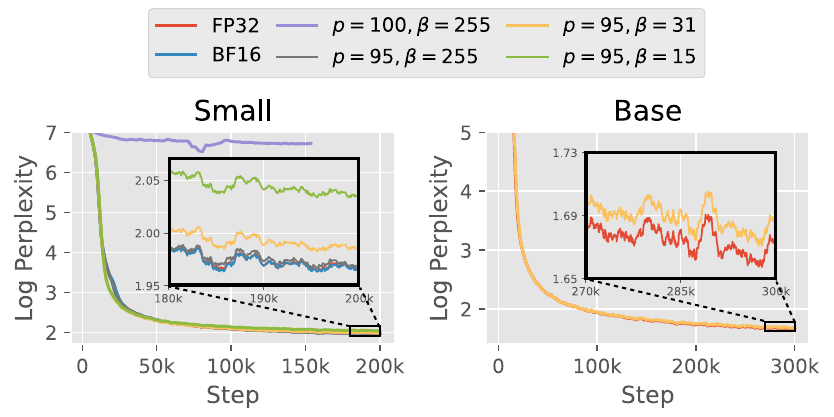}
\vspace{-0.2in}
\caption{Training: Comparison of RoBERTa loss curves. }
\label{fig:roberta_plot}
\vspace{-0.1in}
\end{wrapfigure}

\textbf{Quantize GEMMs in Linear layers.}
% Going beyond memory saving, 
% , especially focus on LLMs, 
It is common \cite{xiao2023smoothquant, llmfp4} to try and quantize the weight and input activation of {\em linear layers} to low precision for compute efficiency. 
We summarize our comparisons in Tab. \ref{tab:quantize_linear}. 
% {\color{red} (just remove?) Taking {\color{red}Vision Transformers} as an example, most results quantize all GEMMs including linear layers and self-attention computation, so we leave the comparison for later. }
Here, we compare RTN to \citep{xiao2023smoothquant, dettmers2022gptint, liu2023llmqat, llmfp4}. As shown in Tab. \ref{tab:quantize_linear}, a simple RTN works remarkably well compared to other baselines. 
% It is able to represent 95\% of values of matrices in GEMMs to a few ($\beta$) distinct integers while still maintaining good performance. 
We use INT as a data type for RTN here; in \S\ref{sec:imunpack}, we show that we can compute integer GEMMs of any bit-widths using arbitrarily low bit-width GEMMs. 

\textbf{Quantize all GEMMs.}
A more ambitious goal is to quantize {\em every} GEMM in a Transformer model for higher efficiency. 
% {\color{red} (just remove?) Again, using Vision Transformers as an example, many existing proposals try to quantize GEMMs in both the linear layers and the computation of self-attention. }
The comparison results with \citep{lin2022fqvit, ivit, PTQ4ViT_arixv2022, apq-vit, li2023repq} are summarized in Tab. \ref{tab:quantized_matmul}. We can draw a similar conclusion that a simple RTN offers strong performance.

\subsection{Efficacy of Integers: Training}
\label{sec:training}

\begin{wraptable}{R}{0.5\textwidth}
\vspace{-0.2in}
\caption{Training: Validation log perplexity of RoBERTa. }
\label{tab:roberta_val_ppl}
% \vspace{-0.03in}
\begin{center}
\begin{small}
\setlength{\tabcolsep}{3.5pt}
\begin{tabular}{lcccccccc}
\toprule
Size & FP32 & BF16 & $\beta=255$ & $\beta=31$ & $\beta=15$ \\
\midrule
Small & 1.869 & 1.868 & 1.823 & 1.840 & 1.891 \\
Base & 1.611 & - & - & 1.601 & - \\
\bottomrule
\end{tabular}
\end{small}
\end{center}
\vspace{-0.1in}
\end{wraptable}

The transition from FP32 to FP16 and BF16 for GEMMs has doubled the compute efficiency of modern deep learning models. However, far fewer efforts have focused on low precision {\em training} (relative to inference) and this usually requires more sophisticated modifications \citep{wang2018training, wu2018training, zhu2020towards, wortsman2023stable}. In this subsection, we evaluate how well quantizing {\em all} GEMMs using RTN works for training Transformer models. To ensure that the updates can be properly accumulated for the parameters, we use FP32 for storing the parameters and use the quantized version for GEMMs. To limit the amount of compute but still gather strong evidence, we evaluate RTN on RoBERTa \citep{roberta} pretraining using masked language modeling \citep{devlin-etal-2019-bert} on the English Wikipedia corpus \citep{wikidump} and ImageNet classification \citep{imagenet_cvpr09} using ViT \citep{vit} (and see T5-Large \citep{t5} finetuning in \S\ref{ap:more_training_experiments}). All hyperparameters (including random seed) are the same for full-precision and RTN quantized training. See \S\ref{ap:training_details} for more details of training configurations.

\begin{wraptable}{R}{0.5\textwidth}
% \vspace{-0.2in}
\caption{Training: Validation top-1 accuracy of ViT-Small.}
\label{tab:vit_val_accu}
% \vspace{-0.03in}
\begin{center}
\begin{small}
\setlength{\tabcolsep}{4pt}
\begin{tabular}{ccccc}
\toprule
FP32 & FP16 & $\beta=63^\dagger$ & $\beta=31^\dagger$ & $\beta=31^*$ \\
\midrule
78.91 & 79.16 & 78.94 & 79.33 & 79.17 \\
\bottomrule
\end{tabular}
\end{small}
\end{center}
\vspace{-0.1in}
\end{wraptable}

\textbf{RoBERTa.}
As shown in Fig. \ref{fig:roberta_plot}, when $p = 95\%$, for both Small and Base models, the RTN quantized training gives an almost identical log perplexity (loss) curves as FP32 training for $\beta \in \{15, 31, 255\}$. For larger $\beta$, the curve is even closer to the FP32 training curve. We see that $\beta = 31$ already gives a remarkably good result. Surprisingly, despite a marginally higher training log perplexity when using RTN, the validation log perplexity of RTN ($\beta = 31$ and $\beta = 255$) is marginally lower than FP32 and BF16, see Tab. \ref{tab:roberta_val_ppl}.

\begin{wrapfigure}{R}{0.5\textwidth}
\centering
\vspace{-0.1in}
\includegraphics[width=0.5\textwidth]{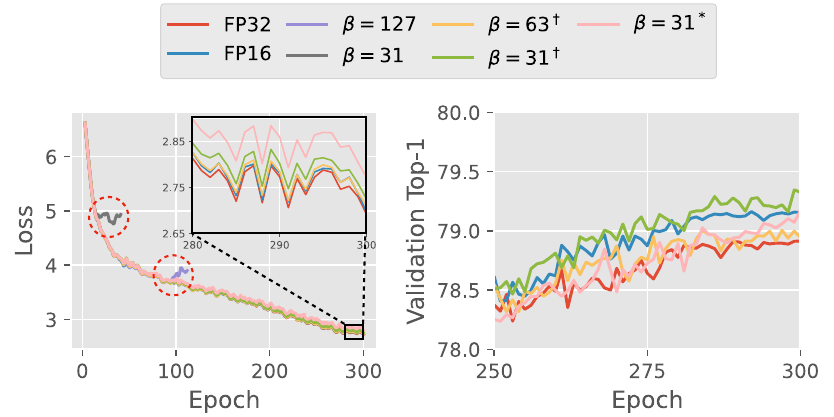}
\vspace{-0.2in}
\caption{Training: Comparison of ViT-Small. $^\dagger$ and $^*$: we set $\beta = 16383$ and $\beta = 1023$, respectively, for the set $\{\nabla_{\mY}, \nabla_{\mP}, \nabla_{\mO}\}$.}
\label{fig:vit_training}
\vspace{-0.1in}
\end{wrapfigure}

\textbf{ViT.}
For ViT, compared to RoBERTa pretraining, we found that it may be necessary to allow the gradients $\nabla_{\mY}, \nabla_{\mP}, \nabla_{\mO}$ of the 
model to have higher bit-width. As shown in Fig. \ref{fig:vit_training}, when $\beta$ is the same ($\beta = 31$ and $\beta = 127$ for the set $\{\mX, \mW, \mQ, \mK, \mM, \mV\}$ and $\{\nabla_{\mY}, \nabla_{\mP}, \nabla_{\mO}\}$, we see divergence in the middle of training. Alternatively, when using larger $\beta$ for only the set $\{\nabla_{\mY}, \nabla_{\mP}, \nabla_{\mO}\}$, the loss curve of RTN quantized training is almost identical to FP32 training. 
Surprisingly, we observed similar results as RoBERTa training: despite marginally higher training loss when using RTN, the validation top-1 accuracy of RTN is higher than FP32 as shown in Fig. \ref{fig:vit_training} and Tab. \ref{tab:vit_val_accu}. 

\section{What happens with Low Bit-Width?}

\label{sec:issue_low_precision_int}

Converting floating point to integers alone will {\em not} provide efficiency benefits. 
Rather, we want to use a representation that can be efficiently computed (and why 
low bit-width integers are common in integer quantization). 
%This is the reason for the universal use of  in integer quantization. 
Notice that as a direct consequence of RTN, by \eqref{eq:quantized_A}, 95\% of values can be represented using $\beta$ distinct numbers, which requires only $\log_2(\beta + 1)$ bits. 
For example, if $\beta = 15$, then we can represent these 95\% of values with 4-bit signed integers, which is already low bit-width. So, is there still a problem? 

It turns out that the difficulty involves dealing with the remaining 5\% of entries. To get a sense of how large these values are, we calculate the ratio $\alpha_{100}(\cdot) / \alpha_{95}(\cdot)$ between the maximum and 95$^{\rm th}$-percentile of the magnitude of each matrix in GEMMs when performing (a) inference (forward pass) of LLaMA-7B and ViT-Large and (b) training (forward pass and backward pass) of RoBERTa-Small at different training phases. We can check the ratios in Tab. \ref{tab:max_percentile_ratio_inference} and Tab. \ref{tab:max_percentile_ratio_roberta_training}, respectively. We see extremely large values across both training and inference and across the entire duration of training, so simply increasing the representation bit width of low precision integers by a few more bits will {\em not} be sufficient to represent these heavy hitters. 

We performed experiments studying different ways of handling these heavy hitters when quantizing all GEMMs (linear layers and self-attention computation) in Transformer models. 
Unless $\beta$ is inordinately large (based on Tab. \ref{tab:max_percentile_ratio_inference} and Tab. \ref{tab:max_percentile_ratio_roberta_training}, more than $10^5$ times larger than our choice of $\beta$ for $p = 95\%$), simply ensuring that the heavy hitters lie within the representable range of $\beta$ for $\beta = 255$ or $\beta = 127$ results in a huge performance drop as shown in Tab. \ref{tab:outliter_issue}. 
%\draftcomment{Vikas}{so the tables/plots so far how are they not showing this?}
% However, when $p = 100$, using extremely large $\beta$ requires more energy, compute, and storage to process high precision arithmetic, and most values only concentrated on very small region of the representable range of high precision integers, resulting in a waste of compute resources. 
On the other hand, clipping the extreme heavy hitters (at the $99.5$-percentile) also fails as shown in Tab. \ref{tab:outliter_issue}. 
% On the other hand, if we just ensure 95\% percent of values are within the representable range of low precision integers ($\beta = 31$ or $\beta = 15$) and let the outliers to be quantized to higher precision integers, the performance could be mostly preserved for inference. 
Our observations for training are similar --  we can see the loss curves for $p = 100\%, \beta = 255$ and $p = 95\%, \beta = 31$ in Fig. \ref{fig:roberta_plot}.

\begin{table}[!tb]
% \vspace{-0.09in}
\caption{Maximal ratios between the maximum and 95-percentile of magnitudes of each matrix involved in GEMMs.}
\vspace{-0.05in}
\label{tab:max_percentile_ratio_inference}
\begin{center}
\begin{small}
\setlength{\tabcolsep}{7pt}
\begin{tabular}{lcccccccc}
\toprule
Model & $\mX$ & $\mW$ & $\mQ$ & $\mK$ & $\mM$ & $\mV$ \\
\midrule
LLaMA-7B & 141312.0 & 47.8 & 8.4 & 8.1 & 4448.0 & 36.2 \\
ViT-Large & 284402.4 & 34.8 & 4.3 & 4.3 & 120.0 & 8.9 \\
\bottomrule
\end{tabular}
\end{small}
\end{center}
\vspace{-0.2in}
\end{table}

\begin{table}[!tb]
% \vspace{-0.09in}
\caption{Maximal ratios between the maximum and 95-percentile of magnitudes of each matrix involved in GEMMs during the training of RoBERTa-Small.}
\vspace{-0.1in}
\label{tab:max_percentile_ratio_roberta_training}
\begin{center}
\begin{small}
\setlength{\tabcolsep}{4pt}
\begin{tabular}{cccccccccc}
\toprule
Progress & $\mX$ & $\mW$ & $\nabla_{\mY}$ & $\mQ$ & $\mK$ & $\nabla_{\mP}$ & $\mM$ & $\mV$ & $\nabla_{\mO}$ \\
\midrule
% 1/6 & 13.0 & 6.0 & 214.0 & 3.6 & 2.7 & 785472.9 & 21146.5 & 3.4 & 21.5 \\
1/3 & 28.7 & 7.1 & 292.5 & 3.7 & 3.0 & 309365.2 & 3924.6 & 3.1 & 25.8 \\
% 3/6 & 29.3 & 12.1 & 283.1 & 4.0 & 2.9 & 331743.5 & 2660.8 & 3.3 & 35.2 \\
2/3 & 25.7 & 13.8 & 235.4 & 4.2 & 2.7 & 283742.8 & 2283.3 & 3.3 & 32.4 \\
% 5/6 & 23.9 & 14.8 & 226.7 & 4.1 & 2.9 & 244406.2 & 2033.1 & 3.4 & 35.2 \\
3/3 & 22.0 & 16.0 & 290.3 & 4.0 & 3.0 & 218376.0 & 2018.6 & 3.4 & 28.9 \\
\bottomrule
\end{tabular}
\end{small}
\end{center}
\vspace{-0.1in}
\end{table}

\begin{table}[!tb]
% \vspace{-0.09in}
\caption{Catastrophic performance degradation when restricting outliers to a representable range of quantized domain or clipping the outliers on zero-shot inference of LLaMA-7B and ImageNet classification of quantized ViT models. $p = 100$ means we keep outliers within representable range of $\beta$ distinct integers. $\beta = \infty$ means that we do not quantize the values. Clip means we clip the values that is larger than $p$-percentile. }
\label{tab:outliter_issue}
\vspace{-0.1in}
\begin{center}
\begin{small}
\setlength{\tabcolsep}{4pt}
\begin{tabular}{cccccccccc}
\toprule
\parbox[t]{3mm}{\multirow{6}{*}{\rotatebox[origin=c]{90}{LLaMA-7B}}} & $p$ & $\beta$ & Clip & ARC-c & ARC-e & BoolQ & HS & PIQA & WG \\
\cmidrule{2-10}
& \multicolumn{3}{c}{Full-Precision} & 43.1 & 76.3 & 77.8 & 57.2 & 78.0 & 68.8 \\
\cmidrule{2-10}
& 100 & 255 & No & 35.8 & 66.2 & 57.8 & 47.4 & 71.3 & 63.9 \\
% 100 & 127 & No & 25.3 & 43.4 & 42.0 & 31.6 & 57.3 & 54.7 \\
& 99.5 & $\infty$ & Yes & 21.4 & 25.5 & 60.2 & 25.8 & 53.5 & 49.9 \\
\cmidrule{2-10}
& 95 & 31 & No & 43.4 & 75.8 & 76.8 & 57.5 & 77.4 & 68.4 \\
% \bottomrule
\cmidrule[0.75pt]{1-10}
\end{tabular}
\setlength{\tabcolsep}{6.3pt}
\begin{tabular}{ccccccccc}
% \toprule
\parbox[t]{3mm}{\multirow{6}{*}{\rotatebox[origin=c]{90}{ViT}}} & $p$ & $\beta$ & Clip & Tiny & Small & Base & Large & Huge \\
\cmidrule{2-9}
& \multicolumn{3}{c}{Full-Precision} & 75.5 & 81.4 & 85.1 & 85.8 & 87.6 \\
\cmidrule{2-9}
& 100 & 127 & No & 53.9 & 69.1 & 72.0 & 81.6 & 83.6 \\
% 99 & $\infty$ & Yes & 2.0 & 5.2 & 0.6 & 4.9 & 0.2 \\
& 99.5 & $\infty$ & Yes & 11.3 & 24.1 & 9.0 & 15.8 & 0.6 \\
\cmidrule{2-9}
& 95 & 15 & No & 71.1 & 79.8 & 84.5 & 85.6 & 87.5 \\
\bottomrule
\end{tabular}
\end{small}
\end{center}
\vspace{-0.05in}
\end{table}

As briefly mentioned earlier, some ideas have been proposed to process these so-called outliers. The approach in \citep{dettmers2022gptint} exploits the location structure of where these outliers occur and moves the columns or rows of each matrix (with these outliers) into a different matrix, then GEMM is performed using FP16. The authors in \citep{xiao2023smoothquant} propose to smooth the outliers in activation and mitigate the quantization difficulty via a transformation. 
This strategy requires specialized GEMM hardware support for different precisions 
and may even lower the performance as shown in our baseline comparisons in \S\ref{sec:inference} and \S\ref{sec:training}. 

{\bf Goals.} We desire an approach that does not alter the results of integer GEMMs; in other words, all results in \S\ref{sec:inference} and \S\ref{sec:training} must remain exactly the same, yet we should not need 
calculations using different precisions. 
This may appear unrealistic but our simple procedure, IM-Unpack, allows representing heavy hitters using low bit-width integers. Calculations are carried out using low bit-width integer arithmetic. 
Specifically, IM-Unpack unpacks a matrix containing heavy hitters into a {\em larger} unpacked matrix (we study how large the expansion will be in \S\ref{sec:unpack_ratio}) whose values are all representable by low bit-width integers. 
IM-Unpack obtains the exact output of the original GEMM using purely low bit-width integer GEMMs on these unpacked matrices.

\section{IM-Unpack: Integer Matrix Unpacking}

\label{sec:imunpack}

Our approach starts with a simple observation that, for example, a 32-bit integer $v$ can be represented as 
\begin{equation}
v = v_0 + 128 v_1 + 128^2 v_2 + 128^3 v_3 + 128^4 v_4
\end{equation}
where $v_i$ are 8-bit integers. Multiplication/addition of two 32-bit integers can be performed on these decomposed 8-bit integers followed by some post-processing steps (scaling via bit shifting and accumulation). 
This unpacking does enable performing high bit-width arithmetic using lower bit-widths, but it achieves this at the cost of requiring more operations. 
For example, one 32-bit addition now becomes five 8-bit additions with some follow up processing, and one 32-bit multiplication becomes twenty five 8-bit multiplications (distributive law). 

\begin{remark}
    The reason why this unpacking is still useful is because the additional work depends on 
    % the size of a workload 
    % which depends on 
    the number and spatial 
    distribution of the heavy hitters/outliers. 
    %spend some additional works on a few workload, but we are able to significantly reduce the overall workload since o
    We harvest gains because outliers account for a very small portion of the matrices that appear in practice in training/inference stages of Transformer models. 
\end{remark}

\begin{wrapfigure}{R}{0.5\textwidth}
\centering
\vspace{-0.25in}
\includegraphics[width=0.5\textwidth]{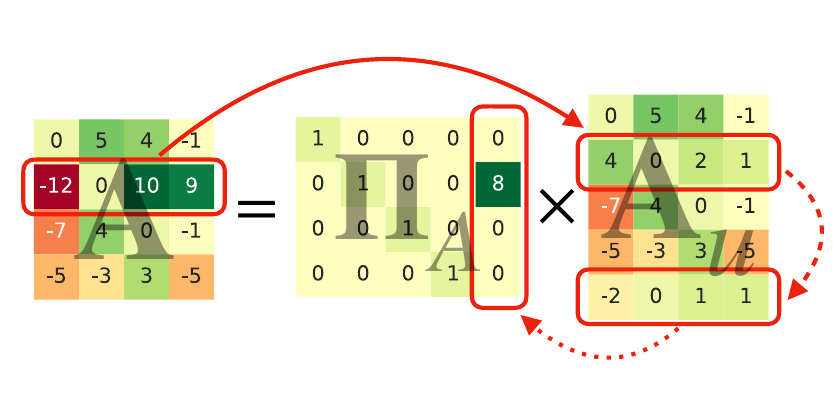}
\vspace{-0.3in}
\caption{Illustration of unpacking row vectors. The solid, dashed, and dotted arrows correspond to lines 5, 4, and 6 in Algo. \ref{alg:unpack_row}}
\label{fig:unpack_row}
\vspace{-0.1in}
\end{wrapfigure}

\setlength{\textfloatsep}{5pt}
\begin{algorithm}[!htbp]
   \caption{UnpackRow($\mA, b$)}
   \label{alg:unpack_row}
   {\small
\begin{algorithmic}[1]
    \STATE Let $\mPi \leftarrow \mI$ and $s \leftarrow 2^{b-1}$ and $i \leftarrow 0$
    \WHILE{$\mA[i, :]$ exists}
        \IF{$\mA[i, :]$ contains OB entries}
            \STATE Append $\text{floor}(\mA[i, :] / s)$ as a new row to $\mA$
            \STATE $\mA[i, :] \leftarrow \mA[i, :] \text{ mod } s$
            \STATE Append $s \mPi[:, i]$ as a new column to $\mPi$
        \ENDIF
        \STATE $i \leftarrow i + 1$
    \ENDWHILE
    \STATE \textbf{return} $\mA, \mPi$
\end{algorithmic}}
\end{algorithm}

Let $b$ be the target bit-width of low bit-width integers and $s = 2^{b - 1}$ be the representable range of bit-width $b$: $b$-bit integers can represent a set $\{-s+1, \cdots, 0, \cdots, s - 1\}$. We refer to any integers inside of this set as In-Bound (IB) values and any integers outside of this set as Out-of-Bound (OB) values, which will be used in later discussion to refer to the values that need to be unpacked. 
We will first show how to unpack a vector to multiple low bit-width vectors. Then, we will discuss how to unpack a matrix using different strategies to achieve better results in different cases. Lastly, we will evaluate how well does IM-Unpack work. 

\textbf{Unpacking an integer vector.} Let $\vv$ be an integer vector and define a function:
\begin{equation}
\begin{split}
m(\vv, s, i) = \text{floor}(\vv / s^i) \text{ mod } s 
\label{eq:mod}
\end{split}
\end{equation}
such that for all $i$, all entries of $m(\vv, s, i)$ are bounded (IB), i.e., lie in the interval $[-s+1, s-1]$. When $s$ is clear from the context, we shorten the LHS of \eqref{eq:mod} to just $m(\vv, i)$. Then,
\begin{equation}
\begin{split}
\vv &= \sum_{i=0}^{\infty} s^i m(\vv, i)
\label{eq:vecter_unpack}
\end{split}
\end{equation}
Note that $\vv / s^i$ decreases to $0$ exponentially fast, so we are able to unpack a vector with just a few low bit-width vectors. 

\subsection{Variants of Matrix Unpacking}

In this subsection, we discuss different strategies of matrix unpacking for different structure-types of matrices. 
First, we discuss the case where $\mA$ is the matrix containing OB values to be unpacked and $\mB$ is a matrix whose values are all IB. Next, we discuss how unpacking works when both $\mA$ and $\mB$ contains OB values. 

\textbf{Unpacking row vectors.} 
We start with the simplest way of unpacking a matrix: unpacking the row vectors. 
Given a matrix $\mA$, if one row of $\mA$ contains OB values, we can unpack the row to multiple rows whose entries are all bounded. The exact procedure is described in Alg. \ref{alg:unpack_row} and illustrated in Fig. \ref{fig:unpack_row}. 
In Fig. \ref{fig:unpack_row}, when the second row in $\mA$ contains OB values, we can unpack it to two row vectors (the second and fifth row) and the post-processing step takes the form of applying $\mPi_A$ to the unpacked matrix $\mA_u$. 

\begin{figure*}[tb]
\centering
\vspace{-0.15in}
\includegraphics[width=0.75\textwidth]{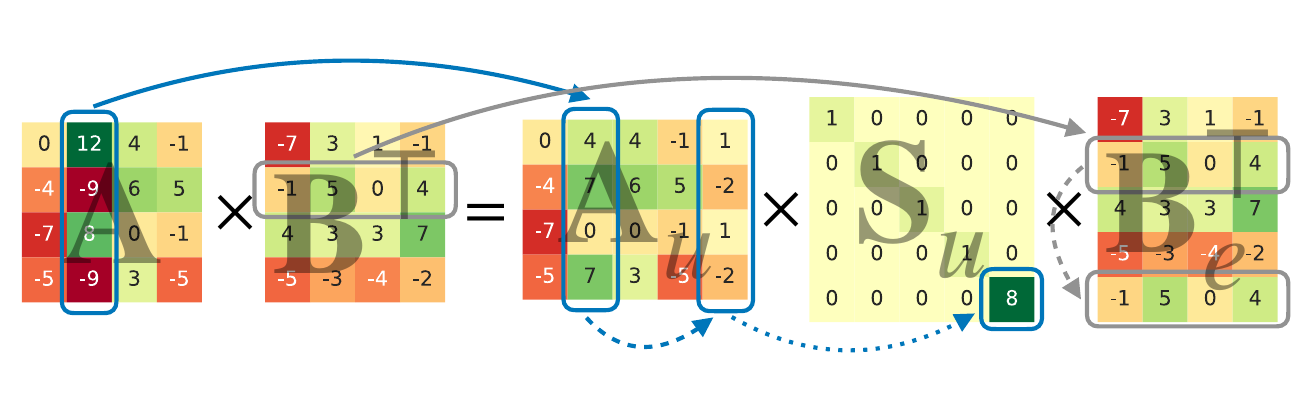}
\vspace{-0.15in}
\caption{Illustration of unpacking column vectors. The blue solid, dashed, and dotted arrows correspond to lines 5, 4, and 7 in Algo. \ref{alg:unpack_row}, and the gray dashed arrow corresponds to line 6 in Algo. \ref{alg:unpack_row}. }
\label{fig:unpack_col}
% \vspace{-0.2in}
\end{figure*}

{\bf Reconstructing $\mA$.} $\mA$ can be reconstructed using the unpacked matrix $\mA_u$ whose entries are IB and a sparse matrix $\mPi$ whose column contains {\em only one} non-zero: 
\begin{equation}
\begin{split}
\mA_u, \mPi_A &= \text{UnpackRow}(\mA, b) \\
\mA &= \mPi_A \mA_u
\end{split}
\end{equation}
Here, applying $\mPi_A$ to $\mA_u$ can be efficiently computed easily (for example, via {\tt \small torch.index\_add}).

% Given another matrix $\mB$ and the corresponding unpacked $\mB_u$ and $\mPi_B$, then multiplying $\mA$ with $\mB^\T$ can be performed via low precision matrix multiplication $\mA_u \mB_u^\T$ followed by the post-processing steps
% \begin{equation}
% \begin{split}
% \mA \mB^\T &= \mPi_A \mA_u \mB_u^\T \mPi_B^\T
% \end{split}
% \end{equation}

\begin{wrapfigure}{R}{0.3\textwidth}
\centering
% \vspace{-0.05in}
\includegraphics[width=0.3\textwidth]{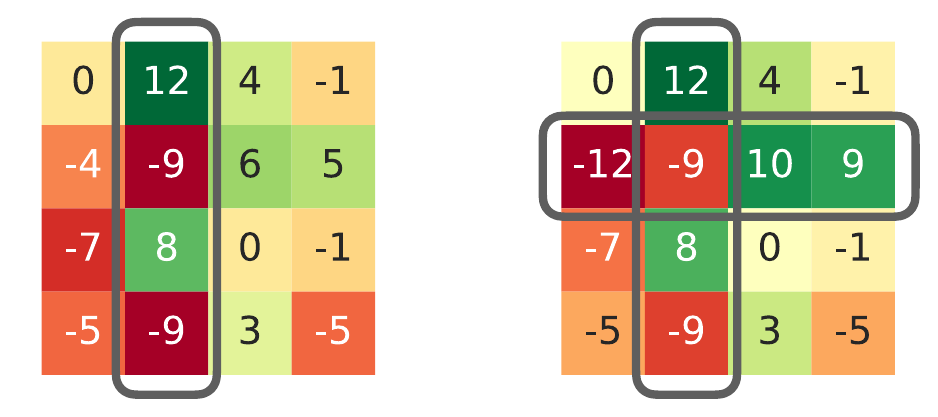}
\vspace{-0.2in}
\caption{Left: Failure case for unpacking rows. Right: Failure case for unpacking rows or columns alone. }
\label{fig:failure_case_unpack}
\vspace{-0.1in}
\end{wrapfigure}

\textbf{Are we done?} If we do not care about maximizing efficiency, then the above scheme already provides a way to perform high bit-width GEMM using low bit-width GEMM. However, this might not be the optimal unpacking strategy for some matrices. For example, consider the left matrix shown in Fig. \ref{fig:failure_case_unpack}. Since every row of this matrix contains OB values, every row need to be unpacked, resulting in a much larger matrix. In this case, it might be better to try and unpack the column vectors. 
Let us apply a similar idea of unpacking row vectors to unpack column vectors of $\mA$:
\begin{equation}
\begin{split}
\mA &= \mA_u' \mPi'_A \\
\mA \mB^\T &= \mA_u' \mPi'_A \mB^\T
\label{eq:unpack_col_problem}
\end{split}
\end{equation}
While unpacking column vectors is reasonable, the sparse matrix $\mPi'_A$ creates an problem. When performing a GEMM of two lower bit-width matrices: 
$\mPi'_A $ has to be applied to $\mA_u'$ or $\mB^\T$ before GEMM, but the result/output may contain OB entries after the application, disabling low bit-width integer GEMM. This problem is similar to  per-channel quantization. 
It is not simple to handle and become more involved when $\mB$ also need to be unpacked.

\textbf{Unpacking column vectors.} Alternatively, let us look at how $\mA \mB^\T$ is computed via outer product of column vectors: 
\begin{equation}
\mC = \mA \mB^\T = \sum_{i = 1}^d \mA[:, i] \mB[:, i]^\T
\end{equation}
Let us look at the $i$-th outer product. Let us try unpacking $\mA[:, i]$ using \eqref{eq:vecter_unpack}, then we have
\begin{equation}
\begin{split}
\mA[:, i] \mB[:, i]^\T &= \sum_{j=0}^{\infty} s^j m(\mA[:, i], j) \mB[:, i]^\T
\end{split}
\end{equation}
Suppose that $m(\mA[:, i], j) = 0$ for $j \geq k$, then we can unpack one outer product to $k$ outer products. This is equivalent to appending $m(\mA[:, i], j)$ for $0 \leq j < k$ to the columns of $\mA$, appending $k$ identical $\mB[:, i]$ to the columns of $\mB$, and maintaining a diagonal matrix to keep track of the scaling factor $s^j$. The exact procedure is described in Alg. \ref{alg:unpack_col}, and Fig. \ref{fig:unpack_col} shows a visualization of unpacking columns. Using column unpacking, we have
\begin{equation}
\begin{split}
\mA_u, \mB_e, \mS_u &= \text{UnpackColumn}(\mA, \mB, \mI, b) \\
\mA \mB^\T &= \mA_u \mS_u \mB_e^\T
\end{split}
\end{equation}
Naively, this still suffers from the same problem as discussed in \eqref{eq:unpack_col_problem} in that there is a diagonal scaling matrix between two low bit-width matrices making low bit-width GEMMs difficult.
% {\color{red} say one sentence more why}. This problem is similar to per-channel quantization. 
However, since $\mS_u$ is a diagonal matrix whose diagonal entries consist of a few distinct factors in $\{1, s, s^2, ...\}$, we can easily compute one GEMM for each distinct diagonal entry as shown in Alg. \ref{alg:scaled_matmul}. 
\begin{equation}
\begin{split}
\mA \mB^\T &= \text{ScaledMatMul}(\mA_u, \mB_e, \mS_u)
\end{split}
\end{equation}
Further, since $s$ is a power of $2$, the scaling can be efficiently implemented via bit shifting. 

\setlength{\textfloatsep}{5pt}
\begin{algorithm}[!htbp]
   \caption{UnpackColumn$(\mA, \mB, \mS, b)$}
   \label{alg:unpack_col}
   {\small
\begin{algorithmic}[1]
    \STATE Let $s \leftarrow 2^{b-1}$ and $i \leftarrow 0$
    \WHILE{$\mA[:, i]$ exists}
        \IF{$\mA[:, i]$ contains OB entries}
            \STATE Append $\text{floor}(\mA[:, i] / s)$ as new column to $\mA$
            \STATE $\mA[:, i] \leftarrow \mA[:, i] \text{ mod } s$
            \STATE Append $\mB[:, i]$ as new column to $\mB$
            \STATE Append $s \mS[i,i]$ as new diagonal entry to $\mS$
        \ENDIF
        \STATE $i \leftarrow i + 1$
    \ENDWHILE
    \STATE \textbf{return} $\mA, \mB, \mS$
\end{algorithmic}}
\end{algorithm}

\setlength{\textfloatsep}{5pt}
\begin{algorithm}[!htbp]
   \caption{ScaledMatMul($\mA, \mB, \mS$)}
   \label{alg:scaled_matmul}
   {\small
\begin{algorithmic}[1]
    \STATE Let $\mC \leftarrow 0$
    \FORALL{distinct diagonal entry $s^i$ in $\mS$}
        \STATE Let $\mathcal{I}$ be the index set where $\mS[j,j] = s^i$ for $j \in \mathcal{I}$
        \STATE $\mC \leftarrow \mC + s^i \mA[:,\mathcal{I}] \mB[:,\mathcal{I}]^\T$
    \ENDFOR
    \STATE \textbf{return} $\mC$
\end{algorithmic}}
\end{algorithm}

% Here, all entries in $\mA_u$ are bounded, but the value range of $\mB_e$ has not been changed. We can further unpack $\mB$ using the same procedure: 
% \begin{equation}
% \begin{split}
% \mB_{eu}, \mA_{ue}, \mathcal{S}_{uu} &= \text{UnpackColumn}(\mB_e, \mA_u, \mathcal{S}_u, b) \\
% \mA \mB &= \mA_{ue} \mathcal{S}_{uu} \mB_{eu}^\T
% \end{split}
% \end{equation}
% Then, the entries in both $\mA_{ue}$ and $\mB_{eu}$ are bounded. 

% \input{figures/failture_case_unpack_row_col}

\begin{figure*}[tb]
\centering
\vspace{-0.25in}
\includegraphics[width=\textwidth]{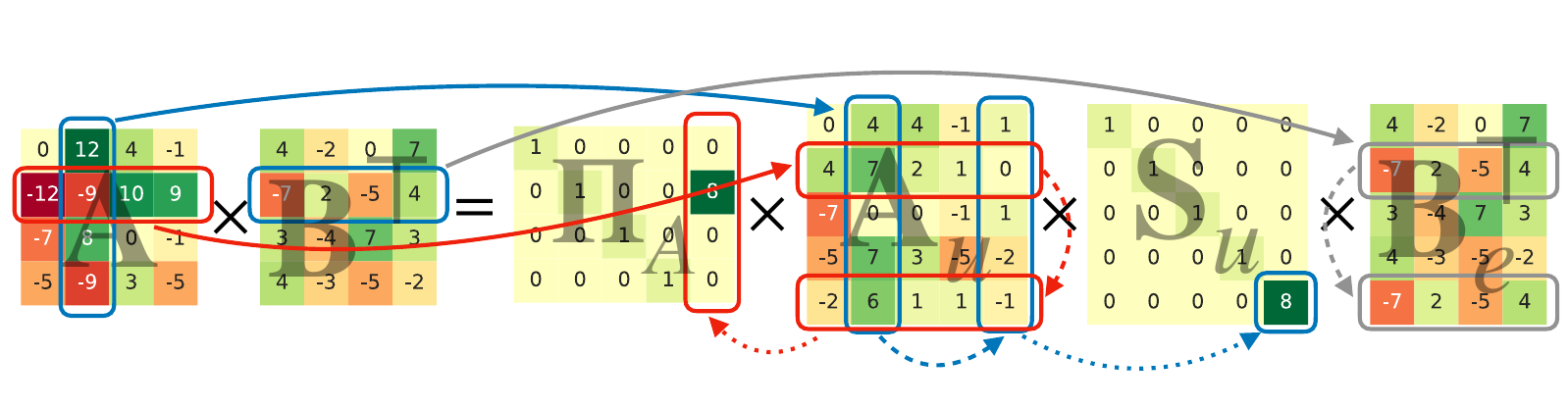}
\vspace{-0.25in}
\caption{Illustration of unpacking both rows and columns based on the OOB counts. The red solid, dashed, and dotted arrows correspond to lines 8, 7, and 9 in Algo. \ref{alg:unpack_both}. The blue solid, dashed, and dotted arrows correspond to lines 12, 11, and 14 in Algo. \ref{alg:unpack_both}, and the gray dashed arrow corresponds to line 13 in Algo. \ref{alg:unpack_both}. }
\label{fig:unpack_both}
% \vspace{-0.2in}
\end{figure*}

\textbf{Are we done yet?} Unpacking columns is efficient for the left matrix shown in Fig. \ref{fig:failure_case_unpack}. However, neither unpacking rows nor unpacking columns will be efficient for unpacking the right matrix shown in Fig. \ref{fig:failure_case_unpack}. All rows and columns contains OB values. Unpacking rows or columns alone will not be ideal. For the right matrix in Fig. \ref{fig:failure_case_unpack}, a better strategy is to unpack the second row and the second column simultaneously.

\setlength{\textfloatsep}{5pt}
\begin{algorithm}[!htbp]
   \caption{UnpackBoth$(\mA, \mB, \mS, b)$}
   \label{alg:unpack_both}
   {\small
\begin{algorithmic}[1]
    \STATE Let $s \leftarrow 2^{b-1}$ and
    \WHILE{True}
        \STATE Let $(c_0, i), (c_1, j)$ be the tuples of top OB count in row/column vectors and corresponding index
        \IF{$c_0 = 0$ and $c_1 = 0$}
            \STATE \textbf{break}
        \ELSIF{$c_0 \geq c_1$}
            \STATE Append $\text{floor}(\mA[i, :] / s)$ as new row to $\mA$
            \STATE $\mA[i, :] \leftarrow \mA[i, :] \text{ mod } s$
            \STATE Append $s \mPi[:, i]$ as new column to $\mPi$
        \ELSE
            \STATE Append $\text{floor}(\mA[:, j] / s)$ as new column to $\mA$
            \STATE $\mA[:, j] \leftarrow \mA[:, j] \text{ mod } s$
            \STATE Append $\mB[:, j]$ as new column to $\mB$
            \STATE Append $s \mS[j,j]$ as new diagonal entry to $\mS$
        \ENDIF
    \ENDWHILE
    \STATE \textbf{return} $\mA, \mB, \mS, \mPi$
\end{algorithmic}}
\end{algorithm}

\textbf{Unpacking both rows and columns simultaneously.} 
Our final strategy combines row and column unpacking together and selectively performs row unpack or column unpack based on the number of OB values that can be eliminated. The procedure is described in Alg. \ref{alg:unpack_both}, and we provide an illustration of unpacking both dimensions in Fig. \ref{fig:unpack_both}. With this procedure, we can obtain the output of high bit-width GEMM using low bit-width as: 
\begin{equation}
\begin{split}
\mA_u, \mB_e, \mS_u, \mPi_A &= \text{UnpackBoth}(\mA, \mB, \mI, b) \\
\mA \mB^\T &= \mPi_A \mA_u \mS_u \mB_e^\T \\
\end{split}
\label{eq:unpack_both}
\end{equation}
Here, $\mA_u \mS_u \mB^\T$ can be calculated via Alg. \ref{alg:scaled_matmul}, and applying $\mPi_A$ can be performed efficiently as discussed.
% (via torch.index\_add for example). 

\textbf{Combining everything.}
Since we have different strategies for unpacking, let us first define a unified interface in Alg. \ref{alg:unpack}. One can verify that for any strategies $s_A$: 
\begin{equation}
\begin{split}
\mA_u, \mB_e, \mS_u, \mPi_A &= \text{Unpack}(\mA, \mB, \mI, b, s_A) \\
\mA \mB^\T &= \mPi_A \mA_u \mS_u \mB_e^\T
\label{eq:unified_unpack}
\end{split}
\end{equation}
In the previous discussion, $\mB$ was assumed to have all IB values. When $\mB$ contains OB values, we note that $\mB$ can be unpacked in a similar manner, and the choice of unpacking strategies for $\mB$ is independent of the unpacking strategy for $\mA$. For example, $\mA$ can be unpacked row-wise, while $\mB$ is unpacked column-wise. By taking the unpacked $\mA_u, \mB_e, \mS_u, \mPi_A$ from  \eqref{eq:unified_unpack}, we can further unpack $\mB$ using strategy $s_B$: 
\begin{equation}
\begin{split}
\mB_{eu}, \mA_{ue}, \mS_{uu}, \mPi_B &= \text{Unpack}(\mB_e, \mA_u, \mS_u, b, s_B) \\
\mA \mB^\T &= \mPi_A \mA_{ue} \mS_{uu} \mB_{eu}^\T \mPi_B^\T
\end{split}
\end{equation}
Here, values in both $\mA_{ue}$ and $\mB_{eu}$ are IB, and the result can be obtained similar to discussion in Eq. \eqref{eq:unpack_both}. 

{\bf Summary.} We introduced three strategies to unpack a matrix to low bit-width integer matrices for different structures of OB values in a matrix. While these strategies work for arbitrary matrices, we can clearly see that these unpacking strategies are most efficient when the OB values concentrate in a few columns and rows. Luckily, the matrices of interest in Transformer models indeed have this property, which is studied and exploited in several works \citep{dettmers2022gptint, xiao2023smoothquant}. 

\setlength{\textfloatsep}{5pt}
\begin{algorithm}[!htbp]
   \caption{Unpack$(\mA, \mB, \mS, b, \text{strategy})$}
   \label{alg:unpack}
   {\small
\begin{algorithmic}[1]
    \IF{strategy is UnpackRow}
    \STATE $\mA_u, \mPi_A \leftarrow \text{UnpackRow}(\mA, b)$
    \STATE $\mS_u, \mB_e \leftarrow \mS, \mB$
    \ELSIF{strategy is UnpackColumn}
    \STATE $\mA_u, \mB_e, \mS_u \leftarrow \text{UnpackColumn}(\mA, \mB, \mS, b)$
    \STATE $\mPi_A \leftarrow \mI$
    \ELSE
    \STATE $\mA_u, \mB_e, \mS_u, \mPi_A \leftarrow \text{UnpackBoth}(\mA, \mB, \mS, b)$
    \ENDIF
    \STATE \textbf{return} $\mA_u, \mB_e, \mS_u, \mPi_A$
\end{algorithmic}}
\end{algorithm}

\subsection{Evaluating Unpacking Overhead}
\label{sec:unpack_ratio}

\begin{table}[!bt]
% \vspace{-0.09in}
\caption{Averaged unpack ratios of each type of GEMMs in LLaMA-7B: linear layers (computing $\mY$), attention score (computing $\mP$), and attention output (computing $\mO$) when using different unpack strategies and integer bit-width $b$ under quantization $\beta$ settings. AS: Attention Score, AO: Attention Output. }
\label{tab:unpack_ratio_llama}
\vspace{-0.1in}
\begin{center}
\begin{small}
\setlength{\tabcolsep}{2pt}
\begin{tabular}{c|cl|cl|ccc|ccc|ccc}
\toprule
\multicolumn{5}{c|}{$\beta$} & \multicolumn{3}{c|}{5} & \multicolumn{3}{c|}{15} & \multicolumn{3}{c}{31} \\
\midrule
\multicolumn{5}{c|}{Integer Bits $b$} & 3 & 4 & 5 & 4 & 5 & 6 & 5 & 6 & 7 \\
\midrule
\parbox[t]{3mm}{\multirow{7}{*}{\rotatebox[origin=c]{90}{Linear ($\mY$)}}} & \parbox[t]{4mm}{\multirow{6}{*}{$\mX$}} & Row & \parbox[t]{4mm}{\multirow{6}{*}{$\mW$}} & Row & 2.67 & 1.93 & 1.57 & 2.47 & 2.02 & 1.73 & 2.12 & 2.00 & 1.74 \\
& & Row & & Col & 10.76 & 2.35 & 1.61 & 9.91 & 5.36 & 1.84 & 8.44 & 5.62 & 1.86 \\
& & Row & & Both & 5.46 & 1.95 & 1.57 & 5.15 & 2.20 & 1.73 & 4.71 & 2.24 & 1.75 \\
& & Col & & Row & 3.80 & 1.32 & 1.06 & 3.98 & 1.64 & 1.16 & 3.93 & 1.68 & 1.17 \\
& & Col & & Col & 15.40 & 1.62 & 1.09 & 16.00 & 4.01 & 1.25 & 15.69 & 4.33 & 1.27 \\
& & Col & & Both & 5.21 & 1.34 & 1.06 & 6.04 & 1.76 & 1.16 & 5.98 & 1.82 & 1.17 \\
\cmidrule{2-14}
& \multicolumn{4}{c|}{Mix} & 2.6 & 1.27 & 1.06 & 2.44 & 1.4 & 1.15 & 2.1 & 1.42 & 1.16 \\
\cmidrule[0.75pt]{1-14}
\parbox[t]{3mm}{\multirow{5}{*}{\rotatebox[origin=c]{90}{AS ($\mP$)}}} & \parbox[t]{4mm}{\multirow{4}{*}{$\mQ$}} & Row & \parbox[t]{4mm}{\multirow{4}{*}{$\mK$}} & Row & 1.97 & 1.60 & 1.0 & 2.00 & 1.87 & 1.15 & 2.00 & 1.87 & 1.18 \\
& & Row & & Col & 3.22 & 1.64 & 1.0 & 5.35 & 2.07 & 1.17 & 5.36 & 2.09 & 1.20 \\
& & Col & & Row & 1.81 & 1.04 & 1.0 & 2.91 & 1.14 & 1.01 & 2.91 & 1.15 & 1.01 \\
& & Col & & Col & 3.36 & 1.08 & 1.0 & 8.66 & 1.32 & 1.03 & 8.67 & 1.35 & 1.03 \\
\cmidrule{2-14}
& \multicolumn{4}{c|}{Mix} & 1.72 & 1.03 & 1.0 & 1.95 & 1.13 & 1.01 & 1.95 & 1.14 & 1.01 \\
\cmidrule[0.75pt]{1-14}
\parbox[t]{3mm}{\multirow{5}{*}{\rotatebox[origin=c]{90}{AO ($\mO$)}}} & \parbox[t]{4mm}{\multirow{4}{*}{$\mM$}} & Row & \parbox[t]{4mm}{\multirow{4}{*}{$\mV$}} & Row & 6.02 & 4.18 & 3.27 & 4.72 & 3.65 & 3.02 & 3.93 & 3.24 & 2.81 \\
& & Row & & Col & 15.10 & 4.53 & 3.35 & 18.21 & 4.64 & 3.16 & 15.07 & 4.21 & 2.95 \\
& & Col & & Row & 16.29 & 8.14 & 5.12 & 11.28 & 6.98 & 4.91 & 8.41 & 5.84 & 4.42 \\
& & Col & & Col & 42.21 & 8.76 & 5.21 & 43.57 & 9.11 & 5.09 & 32.31 & 7.74 & 4.61 \\
\cmidrule{2-14}
& \multicolumn{4}{c|}{Mix} & 5.98 & 4.11 & 3.16 & 4.7 & 3.62 & 2.97 & 3.92 & 3.22 & 2.77 \\
\bottomrule
\end{tabular}
\end{small}
\end{center}
\vspace{-0.1in}
\end{table}

\begin{table}[!tb]
% \vspace{-0.09in}
\caption{Averaged unpack ratios of each type of quantized GEMMs in both forward and backward of a RoBERTa-Small when using different integer bit length $b$ at different training phrases of the $\beta = 31$ experiment in Fig. \ref{fig:roberta_plot}. The optimal strategies (Mix as in Tab. \ref{tab:unpack_ratio_llama}) for each GEMM is used. }
\label{tab:unpack_ratio_roberta}
\vspace{-0.1in}
\begin{center}
\begin{small}
\setlength{\tabcolsep}{4pt}
\begin{tabular}{c|c|ccc|ccc|ccc}
\toprule
\multicolumn{2}{c|}{Progress} & \multicolumn{3}{c|}{1/3} & \multicolumn{3}{c|}{2/3} & \multicolumn{3}{c}{3/3} \\
\midrule
\multicolumn{2}{c|}{Integer Bits $b$} & 5 & 6 & 7 & 5 & 6 & 7 & 5 & 6 & 7 \\
\midrule
\parbox[t]{3mm}{\multirow{3}{*}{\rotatebox[origin=c]{90}{Linear}}} & $\mY$ & 2.00 & 1.31 & 1.08 & 2.00 & 1.32 & 1.07 & 2.00 & 1.32 & 1.05 \\
& $\nabla_{\mX}$ & 1.50 & 1.31 & 1.15 & 1.50 & 1.30 & 1.16 & 1.50 & 1.30 & 1.15 \\
& $\nabla_{\mW}$ & 1.98 & 1.25 & 1.04 & 1.98 & 1.25 & 1.03 & 1.98 & 1.25 & 1.03 \\
\midrule
\parbox[t]{3mm}{\multirow{3}{*}{\rotatebox[origin=c]{90}{AS}}} & $\mP$ & 1.66 & 1.04 & 1.00 & 1.42 & 1.05 & 1.0 & 1.40 & 1.04 & 1.00 \\
& $\nabla_{\mQ}$ & 2.22 & 1.90 & 1.71 & 2.22 & 1.91 & 1.7 & 2.24 & 1.92 & 1.71 \\
& $\nabla_{\mK}$ & 1.79 & 1.06 & 1.00 & 1.49 & 1.07 & 1.0 & 1.45 & 1.07 & 1.00 \\
\midrule
\parbox[t]{3mm}{\multirow{3}{*}{\rotatebox[origin=c]{90}{AO}}} & $\mO$ & 3.11 & 2.71 & 2.30 & 3.10 & 2.68 & 2.24 & 3.10 & 2.62 & 2.22 \\
& $\nabla_{\mM}$ & 1.21 & 1.10 & 1.04 & 1.21 & 1.10 & 1.04 & 1.21 & 1.10 & 1.04 \\
& $\nabla_{\mV}$ & 2.88 & 2.52 & 2.12 & 2.87 & 2.48 & 2.10 & 2.86 & 2.41 & 2.09 \\
\bottomrule
\end{tabular}
\end{small}
\end{center}
\vspace{-0.1in}
\end{table}

\begin{table}[!tb]
% \vspace{-0.09in}
\caption{Averaged ratios of quantized GEMMs ($\beta = 15$) in linear layers on ViT-Large when using different strategies and a range of integer bit-widths $b$ to the lowest bit-width possible. }
\label{tab:unpack_ratio_vit_low}
\vspace{-0.1in}
\begin{center}
\begin{small}
\setlength{\tabcolsep}{4.5pt}
\begin{tabular}{cl|cl|ccccccc}
\toprule
\multicolumn{4}{c|}{Integer Bits $b$} & 2 & 3 & 4 & 5 & 6 & 7 \\
\midrule
\parbox[t]{4mm}{\multirow{9}{*}{$\mX$}} & Row & \parbox[t]{4mm}{\multirow{9}{*}{$\mW$}} & Row & 7.24 & 3.80 & 2.63 & 2.22 & 1.54 & 1.43 \\
& Row & & Col & 194.89 & 27.52 & 10.46 & 4.31 & 1.62 & 1.43 \\
& Row & & Both & 85.92 & 13.80 & 6.22 & 2.76 & 1.56 & 1.43 \\
& Col & & Row & 19.27 & 4.85 & 3.06 & 1.46 & 1.25 & 1.12 \\
& Col & & Col & 526.31 & 35.86 & 12.22 & 2.81 & 1.32 & 1.13 \\
& Col & & Both & 27.06 & 13.31 & 7.59 & 1.78 & 1.26 & 1.13 \\
& Both & & Row & 7.62 & 3.39 & 2.58 & 1.64 & 1.42 & 1.33 \\
& Both & & Col & 206.32 & 24.45 & 10.27 & 3.15 & 1.49 & 1.34 \\
& Both & & Both & 79.19 & 11.16 & 6.09 & 2.01 & 1.43 & 1.34 \\
\midrule
\multicolumn{4}{c|}{Mix} & 6.29 & 2.98 & 2.24 & 1.40 & 1.23 & 1.11 \\
\bottomrule
\end{tabular}
\end{small}
\end{center}
\vspace{-0.1in}
\end{table}

The idea of IM-Unpack is to use more low bit-width arithmetic operations to compute a high bit-width operation. As we see in the description of IM-Unpack algorithm, the number of row and column vectors increases, so the unpacked matrices $\mA_{ue}$ and $\mB_{eu}$ will have a larger size compared to $\mA$ and $\mB$, which obviously increases the computational cost of low bit-width GEMMs. In this subsection, we evaluate how much this cost would increase. For two matrices $\mA$ and $\mB$, the complexity of a GEMM is $\mathcal{O}(ndh)$. Similarly, let $n', d'$ be the size of $\mA_{ue}$ and $h'$ be the number of rows of $\mB_{eu}$. The cost of $\mA_{ue} \mS_{uu} \mB_{eu}^\T$ is $\mathcal{O}(n'd'h')$, we can directly measure the unpack ratio
\begin{equation}
r = (n'd'h') / (ndh)
\end{equation}
to understand by how much the cost for low bit-width GEMMs increases. 
We uses LLaMA-7B to study the unpack ratio $r$ when using different unpacking strategies (Tab. \ref{tab:unpack_ratio_llama}). Note that since unpacking both requires keeping track of the OB count in each row and column vector which is not as fast as the other two strategies, we only use it for unpacking parameters $\mW$ for inference since it can be performed once when loading the model. The Mix in Tab. \ref{tab:unpack_ratio_llama} means that for each GEMM, we compare different strategies and choose the optimal strategy that results in the smallest unpack ratio. We note that the unpack ratios of computing $\mY$ and $\mP$ are quite reasonable, but the ratios of computing $\mO$ is larger. This is expected since the large outliers of the self-attention matrix $\mM$ mainly concentrate in the diagonal \citep{beltagy2020longformer}. 
We also study the unpack ratios of each type of quantized GEMMs at different training phases, and show the results of Mix strategy in Tab. \ref{tab:unpack_ratio_roberta}. The ratios stay relatively unchanged as training progresses. Also, we can observe similar high unpack ratio when computing $\mO$ and $\nabla_{\mV}$ since these GEMMs involve self-attention matrix $\mM$. 
Lastly, we verify that we can unpack matrices to arbitrarily low integer matrices (Tab. \ref{tab:unpack_ratio_vit_low}). The 2-bit setting is the lowest bit width that can be used for symmetric signed integers ($\{-1, 0, 1\}$).

\section{Limitations}

To simplify the presentation, we used the simplest RTN quantization, which might not deliver the optimal performance. More sophisticated techniques 
% on quantization and low precision training 
are likely to further improve the results. For example, we may be able to remove the demands of large $\beta$ for the set $\{\nabla_{\mY}, \nabla_{\mP}, \nabla_{\mO}\}$ for ViT training. 
% However, we leave it for future work. 
%{\color{red} (remove?) Also, the scale of training experiments was kept moderate (tested RoBERTa-Small, RoBERTa-Base, ViT-Small, and T5-Large finetuning (\S\ref{ap:more_training_experiments}) to keep the compute costs reasonable. While the fact that RoBERTa-Small, RoBERTa-Base, and T5-Large training work well without any hyperparameter tuning is a positive preliminary result for much larger models, {\color{red}the efficacy of quantized RTN training on larger models is so not clear}. }
The current unpacking strategies cannot handle the self-attention matrix $\mM$ efficiently since the outliers mainly concentrate on the diagonal region rather than rows or columns; this needs further study. 
% Lastly, although we cast all GEMMs in low precision integer GEMMs, there are still other operations such as non-linear activation and normalization uses floating point operations. During training, the parameters and updates are stored as floating point. Although these computation only account for small amount of the overall computation, if we are able to cast every operations to integer based operations, the design of hardware might be simplified to only support for integer based operations. 

% signed vs unsigned 

\section{Conclusion}

\begin{figure*}[!tb]
\centering
% \vspace{-0.05in}
\includegraphics[width=0.99\textwidth]{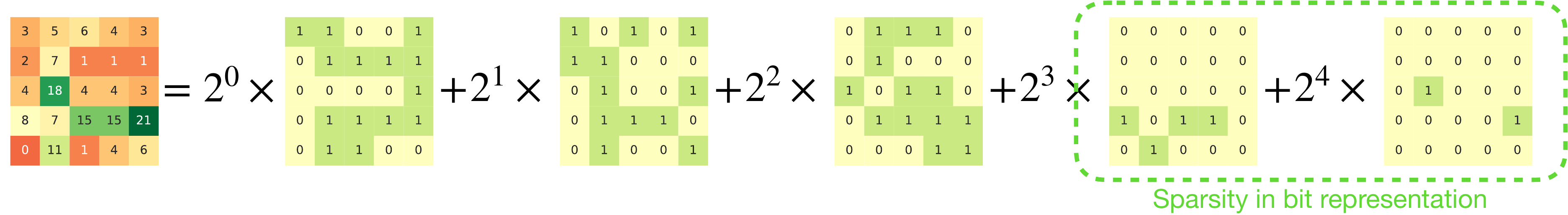}
\vspace{-0.1in}
\caption{Illustration of the bit representation of a matrix. Heavy-hitters have higher order non-zero bits. When a matrix contains heavy-hitters, its bit representation has a sparsity structure in the higher order bits as illustrated. }
\label{fig:sparsity_in_bit}
% \vspace{-0.05in}
\end{figure*}

In this paper, we verify the efficacy of integer GEMMs in both training and inference for Transformer-based models in language modeling and vision. A simple RTN quantization strategy works well compared to baselines. But in this setting, the presence of large outliers/heavy-hitters makes it difficult to make use of efficient low bit-width integer GEMMs since these outliers are much larger than the representable range of low bit-width integers. 
% This is similar is spirit to structured and unstructured sparsity. 
% In a sense, w
We take a ``multi-resolution'' view in how we extract a spectrum of bit-width tradeoffs. This is loosely similar to sparsity but here, instead of 
making a zero versus non-zero distinction between the entries, our heavy-hitters (which need higher bit-width representations) are analogous to ``non-sparse'' entries (as illustrated in Fig. \ref{fig:sparsity_in_bit}). 
% the sparsity is in terms of the bit representation of matrices (note not the bit representation of the tindividual entries). 
%(How well the two paradigms compose together is future work and needs further intellectual and empirical treatment).
To address the challenge of high bit-width heavy-hitters, we develop an algorithm to unpack integer matrices that contains arbitrarily large values to slightly larger matrices with the property that all values lie within the representable range of low bit-width integers and a procedure to obtain the GEMM output of original matrices using only low bit-width integer GEMMs on the unpacked matrices followed by some scaling (using bit shifting) and accumulation. 
Our algorithm can greatly simplify the design of hardware and improve the power efficiency by only supporting low bit-width integer GEMMs for both training and inference.

% \section*{Impact Statement}
% This paper presents an approach to use low precision computing for training and inference. The goal is to accelerate computing speed and reduce energy consumption and thus promote more sustainable computational practices with less environmental concerns. 
% We do not identify any specific societal implications or ethical considerations that must be highlighted. 

\bibliography{main}
\bibliographystyle{plain}

\newpage

\section{Appendices}

In this section, we provide more details about design choices and experiment setups as well as more experiments that are left out in the main text. 

% \subsection{Histogram of All BF16 Values}
% \label{ap:bf_encode}
% \input{figures/bf16-count}

% We plot a histogram of all possible BF16 values in Fig. \ref{fig:bf16_encode}. As we can clearly see that the floating point represents smaller values much more densely compare to large values. 

\subsection{Why Using Percentiles?}
\label{ap:percentiles}

\begin{table}[!htbp]
% \vspace{-0.09in}
\caption{Standard deviation vs percentile when removing largest outliers. $\mX$ has $2.25 \times 10^7$ entries and $\mW$ has $1.68 \times 10^7$ entries. }
\label{tab:percentile_vs_std}
\begin{center}
\begin{small}
\setlength{\tabcolsep}{7pt}
\begin{tabular}{l|l|ccccc}
\toprule
\multicolumn{2}{l|}{Outliers Removed} & $0$ & $10$ & $10^2$ & $10^3$ \\
\midrule
\parbox[t]{4mm}{\multirow{2}{*}{$\mW$}} & Standard Deviation & 0.0082 & 0.0082 & 0.0082 & 0.0082 \\
& 95-Percentile & 0.0177 & 0.0177 & 0.0177 & 0.0177 \\
\midrule
\parbox[t]{4mm}{\multirow{2}{*}{$\mX$}} & Standard Deviation & 0.0330 & 0.0327 & 0.0320 & 0.0214 \\
& 95-Percentile & 0.0280 & 0.0280 & 0.0280 & 0.0278 \\
\bottomrule
\end{tabular}
\end{small}
\end{center}
% \vspace{-0.25in}
\end{table}
We need a way of mapping the actual range of values in a floating point matrix to an integer range, and ensure most values fall into the desired range and fill up the representable range as much as possible, so we need a statistic to estimate the range of values in a FP matrix. 
We compared percentile and standard deviation. We inspected different parameter matrices $\mW$ and the corresponding inputs $\mX$ in the LLaMA-7B model \citep{touvron2023llama}. While the outlier problem in $\mW$ is moderate, and both standard deviation and percentile work well, the outliers in $\mX$ is problematic and contains a few values that are much larger than non-outliers. The estimation of standard deviation might be severely impact the extreme outliers in $\mX$ as shown in Tab. \ref{tab:percentile_vs_std}: removing an extremely small subset of the largest outliers can severely alter the estimates. On contrast, percentile is more robust to the extreme outliers. As a result, we choose percentile as the estimation of value range.

\subsection{Baseline Comparison when Quantize Parameters Only}
\label{ap:quantize_parameters}

\begin{table}[!htbp]
% \vspace{-0.09in}
\caption{Baseline comparison on LLaMA-7B and ViTs when only quantize parameters. HS: HellaSwag, WG: WinoGrande}
\label{tab:quantize_weight}
% \vspace{-0.03in}
\begin{center}
\begin{small}
\setlength{\tabcolsep}{4.28pt}
\begin{tabular}{clllcccccc}
\toprule
\parbox[t]{3mm}{\multirow{10}{*}{\rotatebox[origin=c]{90}{LLaMA-7B}}} & Method & $\beta$ & $\overline{\text{Bits}}$ & ARC-c & ARC-e & BoolQ & HS & PIQA & WG \\
\cmidrule{2-10}
& Full-Precision & - & 16 & 43.1 & 76.3 & 77.8 & 57.2 & 78.0 & 68.8 \\
\cmidrule{2-10}
& GPTQ & - & 4 & 37.4 & 72.7 & 73.3 & 54.9 & 77.9 & 67.9 \\
& LLM-FP4 & - & 4 & 40.4 & 74.9 & 74.2 & 55.8 & 77.8 & 69.9 \\
& QuIP & - & 2 & 22.3 & 42.8 & 50.3 & 34.0 & 61.8 & 52.6 \\
\cmidrule{2-10}
& RTN+HE & 5 & 2.5 & 39.3 & 72.8 & 69.9 & 53.4 & 74.9 & 66.4 \\
&  & 7 & 2.9 & 42.6 & 73.9 & 72.3 & 55.9 & 77.0 & 67.4 \\
% & 9 & 3.3 & 43.0 & 74.5 & 75.7 & 56.4 & 77.4 & 69.4 \\
&  & 11 & 3.5 & 43.9 & 76.1 & 77.3 & 56.3 & 77.3 & 69.3 \\
% & 13 & 3.8 & 44.2 & 75.8 & 77.4 & 56.3 & 77.8 & 69.0 \\
&  & 15 & 4.0 & 43.0 & 75.7 & 77.5 & 57.0 & 78.0 & 69.2 \\
&  & 31 & 5.0 & 42.7 & 76.1 & 76.1 & 57.3 & 77.3 & 69.3 \\
\cmidrule[0.75pt]{1-10}
\end{tabular}
\setlength{\tabcolsep}{6.6pt}
\begin{tabular}{clllccccc}
% \toprule
\parbox[t]{3mm}{\multirow{10}{*}{\rotatebox[origin=c]{90}{ViT}}} & Method & $\beta$ & $\overline{\text{Bits}}$ & Tiny & Small & Base & Large & Huge \\
\cmidrule{2-9}
& Full-Precision & - & 32 & 75.5 & 81.4 & 85.1 & 85.8 & 87.6 \\
\cmidrule{2-9}
& PTQ4ViT & - & 3 & 18.3 & 36.2 & 21.4 & 81.3 & 78.9 \\
% & GPTQ & - & 2 & 0.4 & 0.4 & 29.3 & 63.1 & 42.6 \\
% & QuIP & - & 2 & 1.4 & 22.0 & 77.5 & 82.2 & 84.6 \\
% & FrameQuant & - & 2.2 & 25.8 & 61.5 & 80.9 & 83.7 & 86.0 \\
\cmidrule{2-9}
& RTN+HE & 3 & 1.8 & 0.5 & 8.3 & 63.6 & 81.9 & 83.3 \\
&  & 5 & 2.4 & 38.2 & 69.0 & 81.1 & 84.9 & 86.7 \\
&  & 7 & 2.9 & 63.6 & 76.7 & 83.6 & 85.4 & 87.2 \\
&  & 15 & 4.0 & 73.4 & 80.5 & 84.8 & 85.7 & 87.6 \\
\bottomrule
\end{tabular}
\end{small}
\end{center}
% \vspace{-0.25in}
\end{table}
One direction of quantization research focus on quantizing the parameters for better storage and memory usage. 
We also evaluate how well RTN works on storage and memory efficiency. 
After quantization, the quantized $\mW_q$ usually contains a few hundreds of distinct integers. Simply representing $\mW_q$ in plain integer format would not be efficient and usually requires larger than 8 bits per value for memory. By inspecting the value distribution of $\mW_q$, we found that the fewer values occur much more frequently than others, which create a clear opportunity for compression. We simply apply Huffman Encoding (HE), which was also in \citep{huffman} to compress models for memory efficiency, to use shorter encoding for more frequent values. As shown in Table \ref{tab:quantize_weight}, with RTN and HE, we are able to significantly reduce the average bites per value with small or no performance degradation and result in significantly better efficiency compared to baselines \citep{frantar2023optq, chee2023quip, llmfp4, PTQ4ViT_arixv2022} for both Transformer based LLMs and Vision Transformers. 

\subsection{Details of Training Experiments}
\label{ap:training_details}

We run all of our experiments on NVIDIA RTX 3090's. The following are training hyperparameters. 

\textbf{RoBERTa.} The RoBERTa-Small is a 4-layer Transformer encoder whose model dimension is 512, hidden dimension is 2048, and number of heads is 8. 
For Small models, we train each model for 200K steps with batches of $256$ $512$-length sequences. We use an AdamW optimizer with 1e-4 learning rate, 10,000 warm-up steps, 0.01 weight decay, and linear decay. 
For Base models, we train each model for 300K steps with batches of $128$ $512$-length sequences. We use an AdamW optimizer with 5e-5 learning rate, 10,000 warm-up steps, 0.01 weight decay, and linear decay. 

\textbf{ViT.} We use timm to train our ViT-Small models. The hyperparameters of all experiments are the same: batch size 1024, optimizer AdamW, learning rate 0.001, weight decay 0.05, augmentation rand-m9-mstd0.5-inc1, mixup 0.8, cutmix 1.0.

\subsection{Unpack Ratios of ViT-Large}
\label{ap:unpack_ratio_vit}
\begin{table}[!htbp]
% \vspace{-0.25in}
\caption{Averaged unpack ratios of each type of GEMMs in ViT-Large: linear layers (computing $\mY$), attention score (computing $\mP$), and attention output (computing $\mO$) when using different unpack strategies and integer bit length $b$ under quantization $\beta$ settings. AS: Attention Score, AO: Attention Output. }
\label{tab:unpack_ratio_vit}
% \vspace{-0.03in}
\begin{center}
\begin{small}
\begin{tabular}{c|cl|cl|ccc|ccc|ccc}
\toprule
\multicolumn{5}{c|}{$\beta$} & \multicolumn{3}{c|}{5} & \multicolumn{3}{c|}{7} & \multicolumn{3}{c}{15} \\
\midrule
\multicolumn{5}{c|}{Integer Bits $b$} & 3 & 4 & 5 & 3 & 4 & 5 & 4 & 5 & 6 \\
\midrule
\parbox[t]{3mm}{\multirow{7}{*}{\rotatebox[origin=c]{90}{Linear ($\mY$)}}} & \parbox[t]{4mm}{\multirow{6}{*}{$\mX$}} & Row & \parbox[t]{4mm}{\multirow{6}{*}{$\mW$}} & Row & 2.90 & 2.00 & 1.55 & 3.01 & 2.38 & 1.59 & 2.63 & 2.22 & 1.54 \\
& & Row & & Col & 10.97 & 2.32 & 1.56 & 12.34 & 4.12 & 1.65 & 10.46 & 4.31 & 1.62 \\
& & Row & & Both & 6.24 & 2.08 & 1.55 & 6.84 & 2.82 & 1.60 & 6.22 & 2.76 & 1.56 \\
& & Col & & Row & 2.33 & 1.39 & 1.20 & 3.38 & 1.51 & 1.26 & 3.06 & 1.46 & 1.25 \\
& & Col & & Col & 8.89 & 1.64 & 1.22 & 13.97 & 2.63 & 1.32 & 12.22 & 2.81 & 1.32 \\
& & Col & & Both & 4.99 & 1.44 & 1.20 & 7.99 & 1.76 & 1.27 & 7.59 & 1.78 & 1.26 \\
\cmidrule{2-14}
& \multicolumn{4}{c|}{Mix} & 2.60 & 1.27 & 1.06 & 2.44 & 1.40 & 1.15 & 2.10 & 1.42 & 1.16 \\
\cmidrule[0.75pt]{1-14}
\parbox[t]{3mm}{\multirow{5}{*}{\rotatebox[origin=c]{90}{AS ($\mP$)}}} & \parbox[t]{4mm}{\multirow{4}{*}{$\mQ$}} & Row & \parbox[t]{4mm}{\multirow{4}{*}{$\mK$}} & Row & 1.84 & 1.07 & 1.00 & 2.01 & 1.35 & 1.00 & 1.99 & 1.40 & 1.00 \\
& & Row & & Col & 3.06 & 1.07 & 1.00 & 6.38 & 1.39 & 1.00 & 6.34 & 1.46 & 1.00 \\
& & Col & & Row & 1.34 & 1.01 & 1.00 & 2.50 & 1.04 & 1.00 & 2.49 & 1.05 & 1.00 \\
& & Col & & Col & 2.39 & 1.01 & 1.00 & 8.25 & 1.08 & 1.00 & 8.24 & 1.10 & 1.00 \\
\cmidrule{2-14}
& \multicolumn{4}{c|}{Mix} & 1.33 & 1.01 & 1.00 & 1.91 & 1.04 & 1.00 & 1.90 & 1.04 & 1.00 \\
\cmidrule[0.75pt]{1-14}
\parbox[t]{3mm}{\multirow{5}{*}{\rotatebox[origin=c]{90}{AO ($\mO$)}}} & \parbox[t]{4mm}{\multirow{4}{*}{$\mM$}} & Row & \parbox[t]{4mm}{\multirow{4}{*}{$\mV$}} & Row & 2.84 & 2.07 & 1.65 & 3.07 & 2.24 & 1.80 & 2.56 & 2.11 & 1.78 \\
& & Row & & Col & 5.78 & 2.12 & 1.65 & 11.12 & 2.47 & 1.81 & 9.22 & 2.35 & 1.79 \\
& & Col & & Row & 3.98 & 2.26 & 1.64 & 4.69 & 2.57 & 1.81 & 3.58 & 2.33 & 1.77 \\
& & Col & & Col & 8.42 & 2.29 & 1.64 & 16.97 & 2.83 & 1.81 & 12.92 & 2.60 & 1.77 \\
\cmidrule{2-14}
& \multicolumn{4}{c|}{Mix} & 2.25 & 1.61 & 1.32 & 2.55 & 1.77 & 1.42 & 2.22 & 1.70 & 1.42 \\
\bottomrule
\end{tabular}
\end{small}
\end{center}
% \vspace{-0.1in}
\end{table}

Similar to Tab. \ref{tab:unpack_ratio_llama} in the main text, we also evaluate the unpack ratios of ViT-Large, which are shown in Tab. \ref{tab:unpack_ratio_vit}. The overall results are similar to what was observed in unpack ratios of LLaMA-7B (Tab. \ref{tab:unpack_ratio_llama}). 

\subsection{More Empirical Results on LLM Quantization}
\label{ap:more_inference_experiments}

\begin{table}[!htbp]
% \vspace{-0.03in}
\caption{Baseline comparison on LLaMA-13B when quantize computation in all linear layers.}
\label{tab:llama13b-linear}
\begin{center}
\begin{small}
\setlength{\tabcolsep}{2pt}
\begin{tabular}{lcccccccc}
\toprule
Method & $\beta$ & Type & ARC-c & ARC-e & BoolQ & HellaSwag & PIQA & WinoGrande \\
\midrule
Full-Precision & - & BF16 & 48.0 & 79.5 & 80.6 & 60.0 & 79.2 & 72.1 \\
\midrule
SmoothQuant & - & INT8 & 45.5 & 76.3 & 76.5 & 58.0 & 78.0 & 72.1 \\
 & - & INT4 & 25.1 & 49.9 & 57.6 & 56.0 & 61.3 & 52.6 \\
LLM-FP4 & - & FP4 & 39.9 & 71.7 & 71.9 & 53.3 & 74.8 & 66.7 \\
\midrule
RTN & 5 & INT & 37.6 & 70.0 & 69.1 & 51.9 & 72.4 & 64.6 \\
& 7 & INT & 44.1 & 76.1 & 73.5 & 57.3 & 76.7 & 67.6 \\
& 15 & INT & 46.9 & 78.8 & 79.4 & 59.0 & 78.2 & 72.5 \\
& 31 & INT & 48.0 & 79.7 & 80.2 & 59.9 & 78.0 & 71.3 \\
\bottomrule
\end{tabular}
\end{small}
\end{center}
% \vspace{-0.05in}
\end{table}

\begin{table}[!htbp]
% \vspace{-0.03in}
\caption{Baseline comparison on LLaMA-13B when quantize all GEMMs in a Transformer.}
\label{tab:llama13b-matmul}
\begin{center}
\begin{small}
\setlength{\tabcolsep}{2pt}
\begin{tabular}{lcccccccc}
\toprule
Method & $\beta$ & Type & ARC-c & ARC-e & BoolQ & HellaSwag & PIQA & WinoGrande \\
\midrule
Full-Precision & - & BF16 & 48.0 & 79.5 & 80.6 & 60.0 & 79.2 & 72.1 \\
\midrule
RTN & 5 & INT & 25.1 & 44.4 & 54.8 & 37.7 & 57.9 & 52.0 \\
& 7 & INT & 38.0 & 66.9 & 70.1 & 53.3 & 72.5 & 64.2 \\
& 15 & INT & 45.9 & 77.6 & 80.0 & 59.5 & 77.5 & 71.5 \\
& 31 & INT & 47.9 & 79.3 & 80.0 & 60.5 & 78.6 & 70.9 \\
\bottomrule
\end{tabular}
\end{small}
\end{center}
% \vspace{-0.05in}
\end{table}

\begin{table}[!htbp]
% \vspace{-0.03in}
\caption{RTN performance on Mistral-7B and Phi-2 when quantize computation in all linear layers. }
\label{tab:more_llms}
\begin{center}
\begin{small}
\setlength{\tabcolsep}{2pt}
\begin{tabular}{clcccccccc}
\toprule
& Method & $\beta$ & ARC-c & ARC-e & BoolQ & HellaSwag & PIQA & WinoGrande \\
\midrule
\parbox[t]{3mm}{\multirow{5}{*}{\rotatebox[origin=c]{90}{Mistral-7B}}} & Full-Precision & - & 50.3 & 80.9 & 83.6 & 61.3 & 80.7 & 73.8 \\
\cmidrule{2-9}
& RTN & 5 & 38.1 & 70.5 & 69.9 & 53.9 & 73.3 & 61.4 \\
& & 7 & 44.9 & 75.0 & 76.0 & 58.7 & 77.8 & 68.6 \\
& & 15 & 48.8 & 79.7 & 80.3 & 60.8 & 79.6 & 73.2 \\
& & 31 & 50.3 & 80.1 & 83.5 & 61.5 & 80.7 & 74.4 \\
\cmidrule[0.75pt]{1-9}
\parbox[t]{3mm}{\multirow{5}{*}{\rotatebox[origin=c]{90}{Phi-2}}} & Full-Precision & - & 20.6 & 26.1 & 41.3 & 25.8 & 54.3 & 49.3 \\
\cmidrule{2-9}
& RTN & 5 & 22.1 & 26.7 & 41.5 & 25.6 & 52.3 & 48.1 \\
& & 7 & 21.3 & 25.8 & 40.9 & 25.8 & 53.9 & 49.5 \\
& & 15 & 21.3 & 27.3 & 45.4 & 25.8 & 53.4 & 48.8 \\
& & 31 & 21.0 & 25.8 & 40.8 & 25.7 & 53.0 & 50.7 \\
\bottomrule
\end{tabular}
\end{small}
\end{center}
% \vspace{-0.05in}
\end{table}

To evaluate how well RTN works on the inference of different models and different model sizes, beside the experiments shown in the main text, we also run experiments on LLaMA-13B \citep{touvron2023llama}, Mistral-7B \citep{jiang2023mistral}, and Phi-2 \citep{phi2}. The results are summarized in Tab. \ref{tab:llama13b-linear}, Tab. \ref{tab:llama13b-matmul}, and Tab. \ref{tab:more_llms}. To minimize code change, we only evaluate the quantization of linear layers, as in many quantization works, for Mistral-7B and and Phi-2.

\subsection{More Empirical Results on Training}
\label{ap:more_training_experiments}

\begin{figure*}[!htbp]
\centering
% \vspace{-0.25in}
\includegraphics[width=0.6\textwidth]{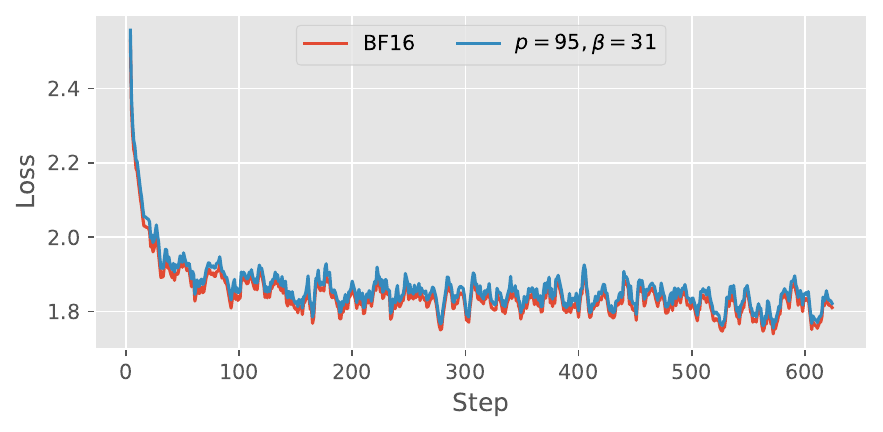}
% \vspace{-0.25in}
\caption{Loss curves of T5-Large finetuning on 1/4 of XSum dataset for 1 epoch. }
\label{fig:t5-large-xsum}
% \vspace{-0.2in}
\end{figure*}

\begin{table}[!htbp]
% \vspace{-0.03in}
\caption{Validation metrics of T5-Large finetuning on 1/4 of XSum dataset for 1 epoch. }
\label{tab:t5-large-xsum}
\begin{center}
\begin{small}
\setlength{\tabcolsep}{2pt}
\begin{tabular}{lcccccccc}
\toprule
Method & $\beta$ & Type & Loss & Rouge1 & Rouge2 & Rougel & Rougelsum \\
\midrule
Full-Precision & - & BF16 & 1.65 & 36.12 & 13.00 & 29.21 & 29.20 \\
\midrule
RTN & 31 & INT & 1.66 & 36.03 & 13.83 & 29.03 & 29.04 \\
\bottomrule
\end{tabular}
\end{small}
\end{center}
% \vspace{-0.05in}
\end{table}

To understand of how well RTN works on the training for larger models without using too much compute, we finetune a T5-Large model on the first 50K instance of XSum summarization dataset \citep{Narayan2018DontGM} using BF16 and RTN, and show the results in Fig. \ref{fig:t5-large-xsum}. The validation metrics are shown in Tab. \ref{tab:t5-large-xsum}. We could draw a similar conclusion as in the main text that RTN quantized training gives similar results as BF16 training.

\end{document}